\documentclass[10pt,twocolumn,letterpaper]{article}

\usepackage[final]{cvpr}              
\usepackage{soul}

\usepackage{graphicx}
\usepackage{amsmath}
\usepackage{amssymb}
\usepackage{booktabs}
\usepackage{times}
\usepackage{epsfig}
\usepackage{float}
\usepackage{algorithm,algpseudocode}
\usepackage{xcolor} 
\usepackage{xspace}
\usepackage{lipsum} 
\usepackage{xcolor}
\usepackage{caption}
\algnewcommand{\LeftComment}[1]{\Statex \(\triangleright\) #1}
\usepackage{mwe}
\usepackage[super]{nth}

\usepackage{authblk}
\usepackage[breaklinks=true,bookmarks=false]{hyperref}

\newcommand{\name}{\textsc{Glass}\xspace}

\makeatletter
\renewcommand\AB@affilsepx{, \protect\Affilfont}
\makeatother
\usepackage[capitalize]{cleveref}
\usepackage[accsupp]{axessibility}
\crefname{section}{Sec.}{Secs.}
\Crefname{section}{Section}{Sections}
\Crefname{table}{Table}{Tables}
\crefname{table}{Tab.}{Tabs.}

\usepackage{xcolor}
\hypersetup{
    colorlinks,
    linkcolor={red!50!black},
    citecolor={blue!50!black},
    urlcolor={blue!80!black}
}

\def\confName{CVPR}
\def\confYear{2022}

\begin{document}

\title{\name: Geometric Latent Augmentation for Shape Spaces}

\author[1]{Sanjeev Muralikrishnan}
\author[2,3]{Siddhartha Chaudhuri}
\author[2]{Noam Aigerman}
\author[2]{Vladimir G. Kim}
\author[2]{Matthew Fisher}
\author[1,2]{Niloy J. Mitra}
\affil[1]{University College London}
\affil[2]{Adobe Research}
\affil[3]{IIT Bombay}

\twocolumn[{%
\renewcommand\twocolumn[1][]{#1}%

\vspace{-2.25cm}
\maketitle
\vspace{-1.25cm}
\begin{center}
    \centering
    \includegraphics[width=\textwidth,height=0.3\textheight]{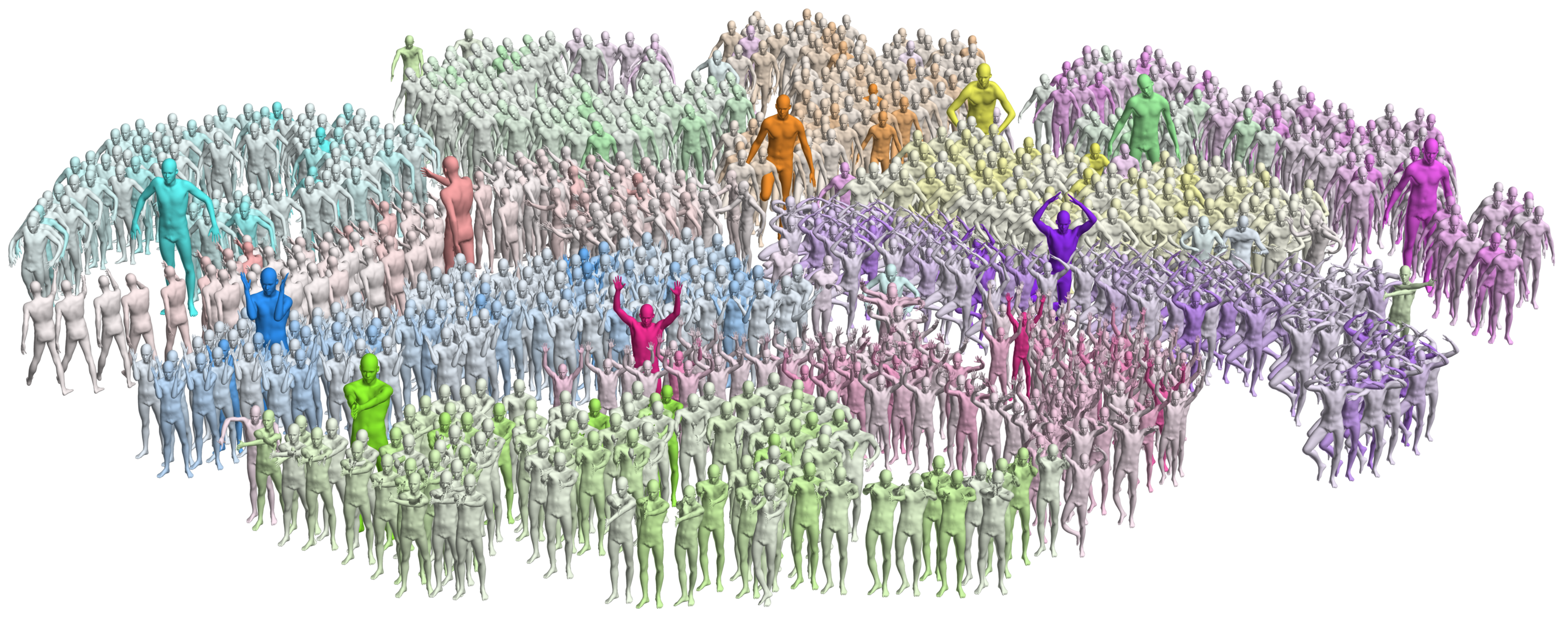}
    \vspace{-7mm}
    \captionof{figure}{Starting from just 10 shapes (larger), our method iteratively augments the collection by alternating between training a VAE, and exploring random perturbations in its low-dimensional latent space guided by a purely geometric deformation energy. Here we show the 1000 most diverse shapes from the first 5K discovered by our method, positioned according to their latent embedding (projected to 2D via t-SNE). Shapes are colored according to the initial landmark they trace back to, with shapes added in later iterations lighter (greyer) in color. The augmentation effectively fills in the space between the sparse initial landmarks, and even extrapolates beyond them. It manages to also interpolate global rotations for samples near the back-facing exemplar, and yields shapes with larger feet-strides (far left), and crossed arms or feet (front, left and center) even though there are no such initial landmarks.
    }
    \label{fig:teaser}
\end{center}%
}]


\begin{abstract}
   We investigate the problem of training generative models on very sparse collections of 3D models. Particularly, instead of using difficult-to-obtain large sets of 3D models, we demonstrate that geometrically-motivated energy functions can be used to effectively augment and boost only a sparse collection of example (training) models. Technically, we analyze the Hessian of the as-rigid-as-possible~(ARAP) energy to adaptively sample from and project to the underlying (local) shape space, and use the augmented dataset to train a variational autoencoder~(VAE). We iterate the process, of building latent spaces of VAE and augmenting the associated dataset, to progressively reveal a richer and more expressive generative space for creating geometrically and semantically valid samples. We  evaluate our method against a set of strong baselines, provide ablation studies, and demonstrate application towards establishing shape correspondences. \name produces multiple interesting and meaningful shape variations even when starting from as few as 3-10 training shapes. Our code is available at \url{https://sanjeevmk.github.io/glass_webpage/}.

\end{abstract}


\section{Introduction}

This paper is concerned with generating plausible deformations of a 3D shape from a very sparse set of examples. Fig.~\ref{fig:teaser} shows an input of 10 human 3D meshes in different poses, and the additional deformations generated by our method.
3D deformations have a strong \emph{semantic} element to them -- e.g., a human's limbs should only bend at the joints, and then, under normal circumstances, not beyond certain angular ranges. Arguably, this can only be deduced in general via learning by example from a dataset.

Unfortunately, in contrast to 2D images, the 3D domain poses several challenges for data-driven frameworks. Probably the most significant one is that data acquisition is complex and tedious, making datasets both scarcer and sparser.

Given this data paucity, we tackle the challenge of generating additional meaningful deformations from a given (very) sparse set of landmark deformations. Our method meaningfully augments the sparse sets to create larger datasets that, in turn, can be leveraged by other techniques that cannot operate on sparse datasets.

Producing plausible deformations from a few landmarks is difficult. Linearly interpolating the vertices of two landmarks yields highly implausible intermediates. A key insight is that while meaningful deformations are semantic, they often have a very strong pure-geometric element, e.g., they are smooth (i.e., preserve local details) and don't distort the shape too much (i.e., local distances are preserved).  However, simply perturbing vertices while minimizing a geometric energy (e.g., smoothness or metric distortion)

generates artifacts such as smooth global bending or surface ripples because by itself, the energy is not a sufficient constraint. Interpolating landmark pairs,  while preserving the energy, fares better but produces limited variations~\cite{alexa2000arapinterp,kilian2007shape}. Our paper, like  other recent approaches~\cite{cosmo2020limp,huang2021arapreg}, advocates {\em learning a low-dimensional generative latent space} which maps out the underlying manifold {\em jointly} defined by the landmarks, while simultaneously minimizing a deformation energy. However, these prior methods still require a large dataset to learn a rich set of variations.

Our core contribution is to address this difficulty with a novel {\em data augmentation} approach that alternates between latent space training and {\em energy-guided exploration}. We employ {\em supervised} learning of a generative space from a training dataset, a very sparse one, but augment that set in an {\em unsupervised}, geometry-aware way. Specifically, we train a Variational Autoencoder (VAE) on the given dataset.

After training, we use the eigenmodes of a deformation energy's Hessian to {\em perturb and project} latent codes of training shapes in a way that ensures they yield smooth, low-distortion deformations, which we add back as data augmentation to the input set. We then re-train the VAE on the augmented dataset and repeat the process iteratively until the space has been densely sampled. In addition to reducing spurious deformations, the use of a low-dimensional, jointly-trained latent space allows low-energy perturbations of one landmark to be influenced by other landmarks, yielding richer variations. We call our method \name.

We evaluate \name on several established datasets and compare performance against baselines using multiple metrics. The experiments show the effectiveness of \name to recover meaningful additional deformations from a mere handful of exemplars. We also evaluate the method in the context of shape correspondence, demonstrating that our sampling process can be used as a data augmentation technique to improve existing strongly-supervised correspondence algorithms (e.g., 3D-CODED~\cite{groueix2018coded}).

\section{Related Work}

\vspace{-1mm}

\paragraph{Geometric shape deformation.} {\em Parametric}
deformation methods express 2D or 3D shapes as a known function of a set of common parameters, and model deformations as variations of these parameters. Such methods include cages~\cite{ju2005meanvalue}, blendshapes~\cite{lewis2014blendshape}, skinned skeletons~\cite{jacobson2014skinning} and Laplacian eigenfunctions~\cite{rong2008spectral}. In contrast, {\em variational} methods model deformations as minimizers of an energy functional -- e.g. Dirichlet~\cite{helein2008harmonic}, isometric~\cite{kilian2007shape}, conformal~\cite{levy2002lscm}, Laplacian~\cite{botsch2008linear}, As-Rigid-As-Possible (ARAP)~\cite{sorkine2007arap}, or As-Consistent-As-Possible (ACAP)~\cite{gao2021acap} -- subject to user constraints. In our work we focus on minimizing the ARAP energy, although our method supports any twice-differentiable energy function. There are strong connections between the parametric and variational approaches, for instance biharmonic skinning weights~\cite{jacobson2011bbw} (parametric) are equivalent to minimizing the Laplacian energy (variational). Please see surveys~\cite{laga2018nonrigid,yuan2021revisit} for a complete discussion. We are also inspired by work on modal analysis~\cite{huang2009modal}, which linearize the deformation space of a shape in terms of the least-significant eigenvectors of the Hessian of some energy functional. In the current paper, we effectively perform {\em learned non-linear modal analysis}: starting with a variational formulation -- the implicitly-defined manifold of low-energy perturbations of a few landmark shapes -- we {\em learn} the corresponding parametric representation as the latent space of an autoencoder by iteratively exploring locally linear perturbations.

Our work on data augmentation from a sparse set of landmark shapes is related to {\em interpolation/morphing} between, and {\em extrapolation} from, sets of shapes. As in our scenario, the set typically comprises articulations of a common template. See e.g.~\cite{tycowicz2015interp} for a survey of classical (non-learning-based) methods for shape interpolation. Plausible extrapolation is less well-defined, and less studied, in the classical literature. Kilian et al.~\cite{kilian2007shape} extend geodesics of an isometric energy in the deformation space, though it is restricted to exploring (and extrapolating) paths between shapes rather than the full deformation space.

\vspace{-4mm}
\paragraph{Learned deformation models.}
Various types of generative models based on graphical models, GANs, VAEs etc have been developed to probabilistically synthesize shape variations. A full treatment is beyond the scope of this paper, please see surveys such as~\cite{chaudhuri2020structgen}. Here, we focus on models which capture the space of smooth deformations of a given shape. The best-studied domain is that of virtual humans, beginning with seminal works capturing face~\cite{blanz1999morph3d}, bodyshape~\cite{allen2003bodyshape} and pose~\cite{anguelov2005scape} variations in a data-driven fashion from scanned exemplars. These works, like several subsequent ones, rely on variations of principal component analysis (PCA) to parameterize the deformation space. Yumer et al.~\cite{yumer2014handles} learn a common set of deformation handles for a dataset. More recent work uses deep neural networks to learn shape deformation models from training sets~\cite{ranjan2018meshae,tan2018meshvae,gao2019sdmnet,gadelha2020handles,wang2019neural}, and use them for applications such as non-rigid correspondences~\cite{groueix2018coded}. Tan et al.~\cite{tan2020thinshell} and Huang et al.~\cite{huang2021arapreg} regularize a VAE with an energy-based loss. We use the latter~\cite{huang2021arapreg} as our choice for energy. However, the primary role of the energy in our method is to guide exploration for data augmentation.

Crucially, all the above methods rely on extensive training data. In contrast, we specifically aim to learn meaningful data-driven deformation models under {\em extreme sparsity constraints}, from just a handful of landmarks indicating modes of the distribution. While this is broadly related to few-shot learning scenarios, only a few other papers consider these requirements in the context of geometric shape synthesis, or without any auxiliary data from other domains. LIMP~\cite{cosmo2020limp} is an important recent work that tries to regularize the latent space of a 3D shape VAE by requiring points sampled on the line segment between two latent codes to minimize geometric distortion relative to the endpoints. Unlike our method, LIMP does not explore the full volume of the hull bounding the training landmarks, or extrapolate beyond it -- regularization is limited to the web of pairwise paths. We modified LIMP to work with ARAP energy, and demonstrate that our method significantly outperforms their approach on a variety of metrics.

\vspace{-4mm}
\paragraph{Unsupervised data augmentation.}
Our work is part of a wide class of methods for synthetically increasing the size of training datasets for data-hungry machine learning, without additional supervision. For broad coverage, we refer the reader to surveys on images~\cite{shorten2019imgaug}, time series~\cite{wen2021timeaug}, and NLP~\cite{chaudhary2020nlpaug}. A particularly relevant recent technique is Deep Markov Chain Monte Carlo~\cite{shahbaba2019deepmcmc}, which samples perturbations of training data using MCMC on an energy functional, trains an autoencoder on these samples, and uses the resulting latent space for lower-dimensional (and hence faster) MCMC. We observed that on very sparse and high-dimensional datasets (only a few landmark 3D shapes), the initial samples of Deep MCMC do not capture meaningful variations, and hence it does not adequately augment the dataset. Also related are methods that augment classification datasets with adversarial perturbations along the gradients of loss functions~\cite{goodfellow2015adversarial,shankar2018crossgrad}. In contrast, we seek to {\em preserve} an energy-based loss, and hence {\em eliminate} the gradient and other high-change directions from consideration.

\section{Method}

\subsection{Problem Setup}
\label{ss:problem}
We assume all shapes in a particular input dataset are meshes with consistent topology. Given a mesh with $N$ vertices  $V\in\mathbb{R}^{N\times 3}$ and triangle faces $T$, a mesh  deformation is simply an assignment of a new position to each vertex, denoted as $W\in\mathbb{R}^{N\times 3}$. We consider the input dataset itself as deformations of a base topology and we are given a sparse set of $n$ deformation ``examples",  $W^1,\dots W^n$. We assume access to a deformation energy $f(W,W')$ which measures the distortion of candidate deformation $W$ with respect to an exemplar deformation $W'$, with higher values indicating more severe distortion induced by the candidate. For brevity, we omit $W'$ and simply write $f(W)$ to mean energy with respect to the relevant base shape. We use the As-Rigid-As-Possible~(ARAP) energy~\cite{sorkine2007arap}  and its latent-space approximation ARAPReg~\cite{huang2021arapreg} to measure the deviation of a deformation from isometry, i.e., how much do geodesic lengths change with respect to the rest pose $V$.

\begin{figure}[t!]
    \centering
    \includegraphics[width=\linewidth]{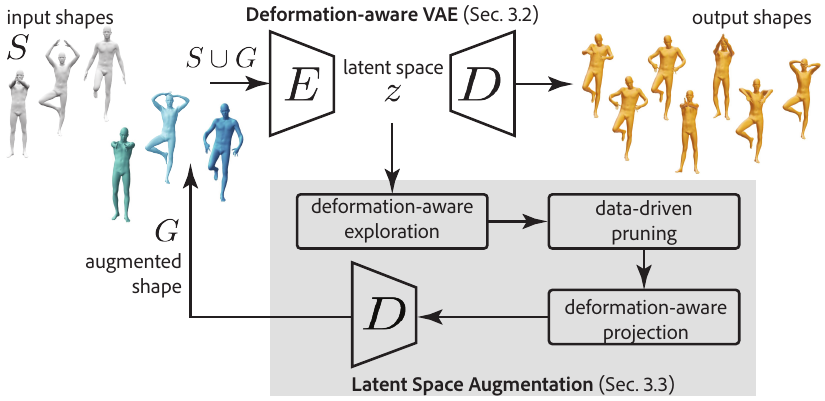}
    \caption{We present \name to iteratively build a deformation-aware VAE latent space and analyzing it to generate new training samples to augment the original training set. This enables generation of diverse yet plausible shape variations starting from very few input examples.}
    \label{fig:overview}
    \vspace{-\baselineskip}
    \vspace{-0.5em}
\end{figure}

We devise a subspace-sampling strategy that adheres to two properties: (i)~it should be data-driven, and contain deformations from the given sparse set; and  (ii)~it should be geometrically-meaningful, i.e., the deformations should have low energy, wrt the given deformation energy $f(W)$.

Our main contribution is a method for online data augmentation during the training of a variational autoencoder (VAE)~\cite{kingma2014autoencoding}. Namely, during training, our method explores the current sample space, guided by the deformation energy $f(W)$, to discover additional meaningful deformation samples. These are progressively used as additional sample points to form an augmented dataset that is used to retrain the VAE, and the process is iterated until convergence.

\subsection{Deformation-Aware VAE}
\label{ss:augmentation}
Let
 $E:~\mathbb{R}^{N\times 3}\to\mathbb{R}^{K}$ be the encoder in the standard VAE architecture, mapping a deformation $W$ into vectors of mean $\mu$ and variance $\Sigma$ into a distribution $E(W) \sim \mathcal{N}(\mu, \Sigma)$. These vectors define the mean and variance of a multivariate Gaussian distribution from which the  latent code $z$, of dimension $K$, is sampled, i.e., $z\sim \mathcal{N}(\mu,\Sigma)$. Similarly, let $D:\mathbb{R}^K\to\mathbb{R}^{N\times 3}$ be the decoder mapping the latent code to a deformation, $D(z) = W$. We shall slightly abuse notation and use $D(E(W))$ to denote the full autoencoding process of $W$, including the step of sampling from the Gaussian. We define three losses to be used in training.

\begin{figure}[b!]
    \centering
    \includegraphics[width=\columnwidth]{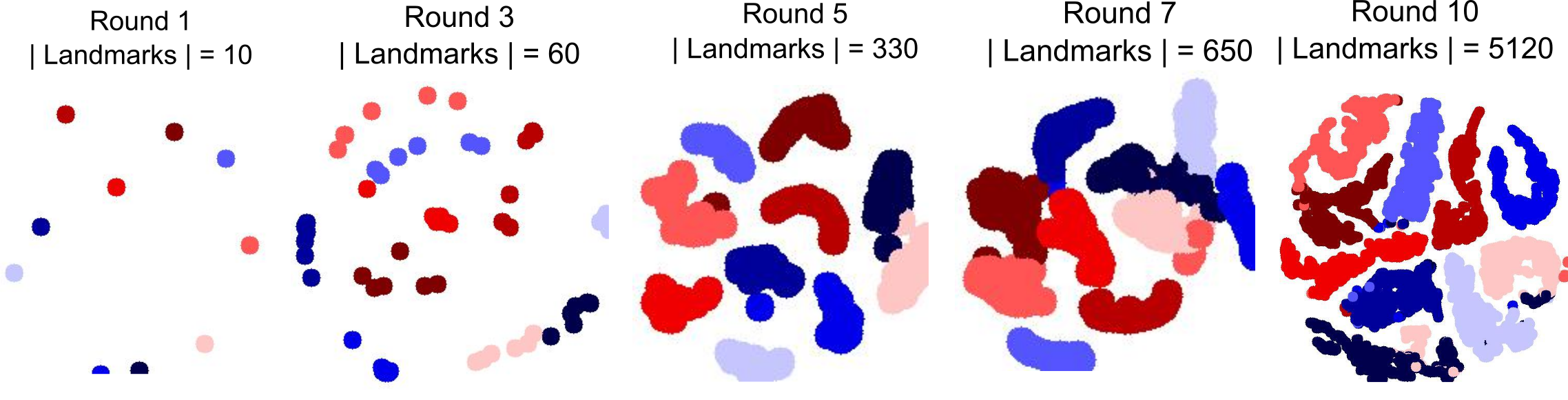}
    \caption{
    tSNE embedding of generated samples shows progressive augmentation of the shape space. Sample color indicates originating (parent) shape. See also Fig.~\ref{fig:teaser}.
    }
    \label{fig:sampling_over_iterations}
    \vspace{-\baselineskip}
\end{figure}

\textit{(i) Reconstruction Loss:}
We require that the VAE reduces to an identity map for any sample deformation,
\vspace{-2mm}
\begin{equation}
  L_\text{Reconstruction} := \lVert D(E(W)) - W \rVert ^ 2.
  \vspace{-2mm}
\end{equation}

\textit{(ii) Gaussian Regularizer Loss:}
Instead of the standard the KL divergence regularizer used in VAEs, we constrain the sample mean and covariance of the mini-batch to that of a unit Gaussian, as proposed in \cite{gaussian_reg}. We found that for a small sample size this batch-based loss leads to faster convergence, compared to the standard KL-Divergence. We denote this loss as,
\vspace{-2mm}
\begin{equation}
  L_\text{Gaussian} := \frac{1}{b} \sum_{i=1}^{b} \left(\lVert \mu_i \rVert ^ 2 + \lVert \Sigma_i - \mathbb{I} \rVert ^ 2\right),
  \vspace{-2mm}
\end{equation}
where $b$ is the mini-batch size, $\mu_i,\Sigma_i$ are the predicted mean and covariance for the $i$-th sample in the mini-batch, and $\mathbb{I}$ is the identity matrix.

\textit{(iii) Deformation Energy:}
Lastly, we require the resulting deformation to have low deformation energy,
\vspace{-2mm}
\begin{equation}
\label{eq:deformation}
    L_\text{Deformation} := f(D(E(W))).
  \vspace{-2mm}
\end{equation}
In summary, our network training loss is
\vspace{-2mm}
\begin{equation} \label{eq:loss}
    L := L_\text{Reconstruction} + L_\text{Gaussian} + \sigma L_\text{Deformation}
  \vspace{-2mm}
\end{equation}
where $\sigma$ is a scalar weight applied to the energy function.

\begin{algorithm}[t!]
\begin{algorithmic}[1]
 \Procedure{\sc LatentAugment}{$W$,$E$,$D$,$f$,$R$}

 \LeftComment{E = Encoder, D= Decoder}
 \LeftComment{W = Initial deformation, f = Energy}
 \LeftComment{R = All previously input or generated shapes}
 \State {$l = E(W)$} \Comment{latent code}
 \State {$\overline{H} \approx \nabla_l \nabla_l f(D(l))$} \Comment{approximate Hessian}
 \State {$\lambda, U^{\uparrow}(H) = \text{EigenDecomposition}(H)$}
 \State {$\lambda_k , U^{\uparrow}_{k}(H) \leftarrow \lambda, U^{\uparrow}(H)$} \Comment{retain k components}
 \State $W_d = \emptyset$
 \For{j $\in$ [1,s]:}
    \State $\beta \sim \mathcal{N}(0,\mathbb{I}) \in R^k$ \Comment{sample $\beta$, $k \ll K$}
    \State $\hat{\beta} = \beta/ \sum_{i=1}^{k} \beta_i^2  $
    \State $\alpha = \sqrt{ 2\delta/\sum_{i=1}^{i=k} \hat{\beta}_{i}^{2} \lambda_i} $
    \State $\hat{W}_j = D(l + \alpha \sum_{i=1}^{k}\hat{\beta}_i U^{\uparrow}_{i}(H))$
    \State $W_d \leftarrow W_d \cup \hat{W}_j$ \Comment{add to candidates}
\EndFor

 \State $W^* = \text{MMR}(W,R,W_d)$ \Comment{prune candidates}
 \State $W_\text{Projected} = \arg\min f(W^*)$ \Comment{project}
 \State $R \leftarrow R \cup \{ W_\text{Projected} \}$ \Comment{augment training set}
 \EndProcedure
 \end{algorithmic}
\caption{Pseudocode for searching the latent space, starting from deformation $W$, for a new augmenting shape.}
\label{alg:proj}
\end{algorithm}

\subsection{Augmenting via Latent Space Exploration}
\label{sec:augmentation}
We now describe the main step of our technique -- adding additional deformation examples $W^j$ to the latent space to reinforce training (Algorithm \ref{alg:proj}).
Simply optimizing (\ref{eq:loss}) is not enough to cover the deformation space. Instead, we continuously introduce new low-energy deformations into the training set, by which we make the \emph{data term aware of the deformation energy}. We achieve this through three steps: (i)~deformation-aware \emph{perturbation} of the latent code in directions that locally \textit{least} modify the deformation energy; (ii)~data-driven \emph{pruning} of perturbed codes that do not introduce variance to the current dataset; and (iii)~deformation-aware \emph{projection} of the new codes to further lower their deformation energy, with an optional \emph{high-resolution projection} to transfer the deformation from a low-resolution mesh to a higher resolution one. Figure ~\ref{fig:sampling_over_iterations} illustrates how the latent space is progressively populated with new deformations over iterations, where colors indicate the base shapes.

\vspace{-3mm}
\paragraph{(i) Deformation-aware perturbation in latent space.} Our goal is to create variations of a given code in latent space without modifying its deformation energy significantly: let $W$ be a deformation, and $l = E(W) \in \mathbb{R}^K$ a latent code achieved from encoding it. We aim to find the low-energy perturbation modes of $l$. In short, we aim to perturb the deformation while not changing the deformation energy too much, or in other words -- we wish to stay on the current level set of the energy. To achieve that, we can restrict ourselves to perturbations on the local \emph{tangent space} of the energy's level set. This tangent space simply comprises of all directions orthogonal to the gradient $\nabla_l f(D(l))$.

In the tangent space, we can pick better directions using a second order analysis. Let $H$ denote the Hessian of the deformation energy with respect to the latent code,
\vspace{-2mm}
\begin{equation} \label{eq:hessian}
    \begin{split}
        H := \nabla_l \nabla_l f(D(l)).
    \end{split}
  \vspace{-2mm}
\end{equation}
Let $\lambda_i$, and $U^{\uparrow}_{i}(H)$ respectively denote the eigenvalues and eigenvectors  of $H$, in ascending order of eigenvalues. Since smaller eigenvalues correspond to directions of less change in energy, we retain only the $k$ ($k << K$) smallest $\lambda_i$  and $U^{\uparrow}_{i}(H)$.
We then draw a random perturbation, by sampling a random vector $\beta \in \mathbb{R}^k$ from a normal distribution. Each $\beta_i \in \beta$ corresponds to step along the eigenvector $U^{\uparrow}_{i}(H)$. We normalize $\beta$ to $\hat{\beta}$ such that $\sum_{i=1}^k {\hat{\beta}_i U^{\uparrow}_{i}(H)} $ is a unit vector, i.e., $\sum_{i=1}^k \hat{\beta}_i^2 = 1$. We then take a step,
\vspace{-2mm}
\begin{equation} \label{eq:update}
    l_t := l + \alpha \sum_{i=1}^{k} {\hat{\beta}_i U^{\uparrow}_{i}(H)},
  \vspace{-2mm}
\end{equation}
where $\alpha$ denotes the step-size and $l_t$ is in the tangent plane. We repeat this process $s$ times for each latent code $l$ 
to get $s$ perturbed codes $\Tilde{l}^1,\Tilde{l}^2,\dots \Tilde{l}^s$. Let $\{\Tilde{W}^1, \Tilde{W}^2\dots \Tilde{W}^s\}$ denote the decoded perturbed deformations where $\Tilde{W}^j = D(\Tilde{l}^j)$.

\textbf{Variable step size:}
Different regions of the latent space have different local deformation landscapes, e.g., curvatures along different directions on the tangent plane. Hence, we should adapt $\alpha$  to the nature of the local landscape around $l$. To that end, we formulate the step size $\alpha$ in terms of the eigenvalues $\lambda_i$ and a user-prescribed threshold $\delta$ on the allowed deformation energy $f$ (i.e., $f()\le \delta$).
Assuming C2 continuity for deformation energy $f()$, we can obtain the following bound on the step size (see supplemental),
\[
    \alpha \le  \sqrt{\frac{2\delta}{\sum_{i=1}^{k} \hat{\beta}_{i}^{2} \lambda_i}}.
\]
This gives an upper bound on the step size along any deformation direction.

\paragraph{(ii) Data-driven pruning of the perturbed deformations.}
In order to add diverse samples, given the set of candidate deformations $\{\Tilde{W}^j\}$, we select one example to be added to the dataset, via   Maximal Marginal Relevance (MMR) ranking~\cite{MMR_1998}.
Specifically, MMR gives a higher score to  perturbations that are   similar to the unperturbed $W$, but different from the existing deformations set $R$, containing the landmarks and deformations generated so far.
We compute as,
\vspace{-1.5mm}
\begin{align}
    \begin{split}
     F(w) = \gamma M(w,W) - (1-\gamma)\max_{r \in R} M(w,r),
    \end{split}
  \vspace{-2mm}
\end{align}
where $M(x,y)$ is the cosine similarity between $x$ and $y$.
Thus, we choose the deformation ${W^* \in W_d}$ that maximizes the MMR function with its latent code  denoted $l^*$.

\vspace{-4mm}
\paragraph{(iii) Deformation-aware projection to smooth, low-energy deformations.}
Although the deformation-aware perturbation somewhat avoids high-energy states, the perturbed deformation $W^*$ may still exhibit undesirable artifacts such as lack of smoothness or high deformation energy. Hence, we project the code to have lower deformation energy \emph{with respect to the unperturbed $W$}. We achieve this by treating $W$ as the rest pose, defining a deformation energy with respect to it, $f_W$. We perform a fixed number of gradient descent steps starting from $W^*$ to lower the energy, which yields the final deformation $W_{\text{Projected}}$. In our experiments, we optimize up to the threshold of $10^{-5}$.

\vspace{-3mm}
\paragraph{(iv) Augment and iterate:} Finally, we append the newly generated deformations to the current (training) set, and continue training. We repeat this augmentation and retraining several times, until we reach a target training set size.

\hspace{-5mm}
\subsection*{Implementation Choices}
\label{ss:implementation}
\noindent \textbf{Choice of energy $f$.} For training the VAE, we set $f$ as the $L2$ formulation of the ARAPReg energy~\cite{huang2021arapreg}. ARAPReg is an efficient approximation of the ARAP energy~\cite{sorkine2007arap} that is conducive for network training. This energy computes an approximate Hessian $\overline{H}$ of ARAP with respect to the latent space, and directly minimizes the eigenvalues of $\overline{H}$.
\newline
\textbf{Approximation of Hessian.} Similarly, for step (i) in Section \ref{sec:augmentation}, we use the approximate Hessian $\overline{H} \approx J^THJ$ proposed in \cite{huang2021arapreg} in our Equations~\ref{eq:hessian} and \ref{eq:update}, where $H$ is the exact Hessian of ARAP and $J$ is the stochastic Jacobian of the ARAP with respect to the $\mathbb{R}^K$ latent space. 
\newline
The above choices speed up training and yield better results than classical ARAP. We still use ARAP for step (iii) where Hessian approximation is not needed.
\newline
\textbf{Upper bound for $\alpha$.} The eigenvalues allow us to choose an appropriate local step-size, but they only yield a local approximation of the deformation energy. As our method continues adding low-energy shapes, the eigenvalues become lower, leading to an extremely large step size $\alpha$. Thus, we set an upper bound of $2$ on $\alpha$.
\newline
\textbf{High-resolution projection.} To speed up training and avoid deformations that are too high-frequency, we use low-resolution meshes (low vertex count). We decimate the high-resolution meshes by preserving a subset of the vertices, thus retaining correspondence from low to high. The generated low-res deformations can later be projected back to original high-res meshes by treating the chosen subset of vertices as deformation constraints in the ARAP optimization proposed in \cite{sorkine2007arap}.

\begin{figure}[t!]
    \centering
    \includegraphics[width=\linewidth]{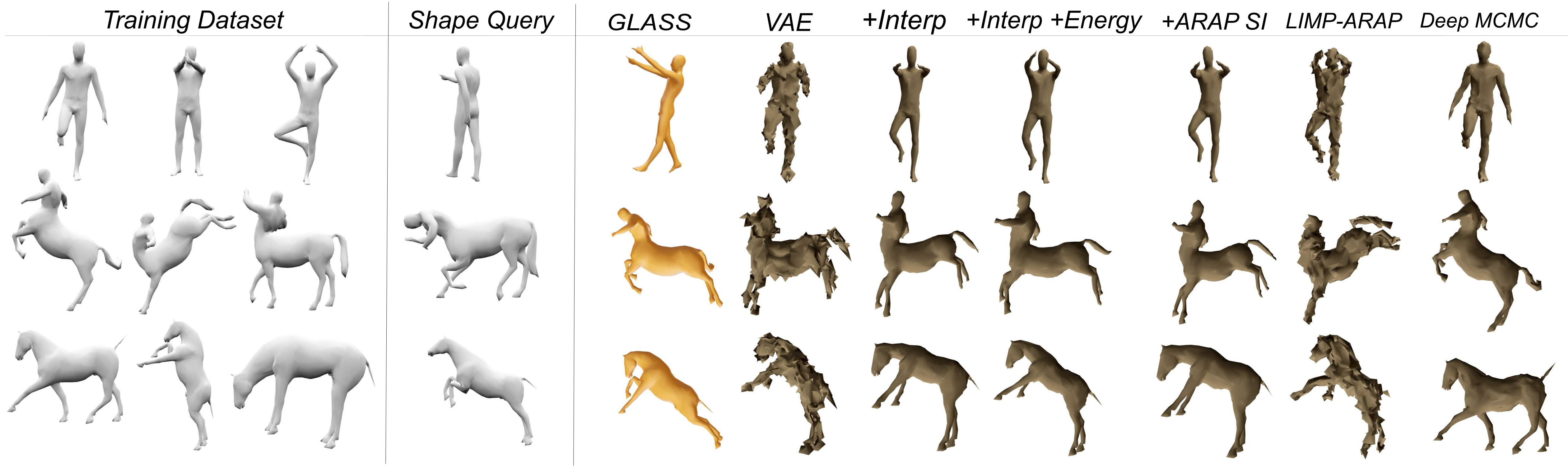}
    \caption{\textit{Generation results evaluated by coverage.} We train different methods on the same training data (col 1) and generate comparable numbers of shapes. Given two shapes from the holdout data (col 2), we evaluate the methods by finding the closest generated shape (cols 3-9). Note how the baselines exhibit strong artifacts and usually do not match the query shape.}
    \label{fig:generation_comparison}
    \vspace{-\baselineskip}
\end{figure}

\begin{figure}[t!]
    \centering
    \includegraphics[width=\columnwidth]{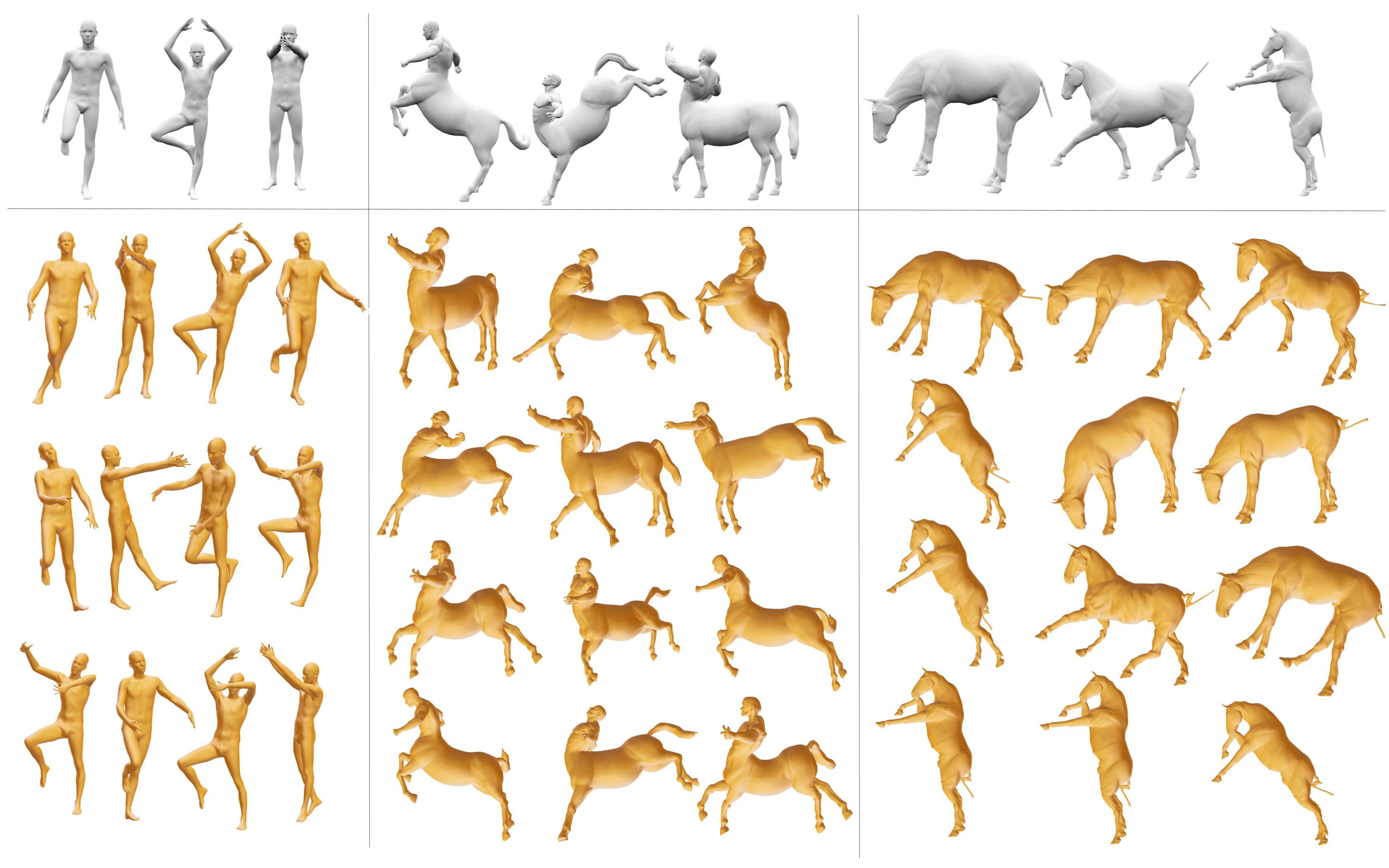} 
    \caption{Training \name on the human, centaur, and horse meshes using the 3 examples each~(top). (Bottom)~We show random samples from the latent space, which combine different properties learned from the example deformations.}
    \label{fig:results_various}
    \vspace{-\baselineskip}
\end{figure}


\section{Experiments}

We evaluate \name on public data of humans (FAUST~\cite{bogo2014faust}) and creatures (TOSCA~\cite{bronstein2008numerical}) in different poses. In our experiments, we sample $X$ landmark poses of a model from a dataset and train our method. We evaluate quality and diversity of newly generated poses as well as interpolation sequences between landmark poses.  We denote our experiments as ``SubjectName-$X$'' to indicate the type of the subject and the number of landmark poses provided as an input to our method; most results use between 3 and 10 landmarks.

\begin{figure}[b!]
    \centering
    \includegraphics[width=1.0\linewidth]{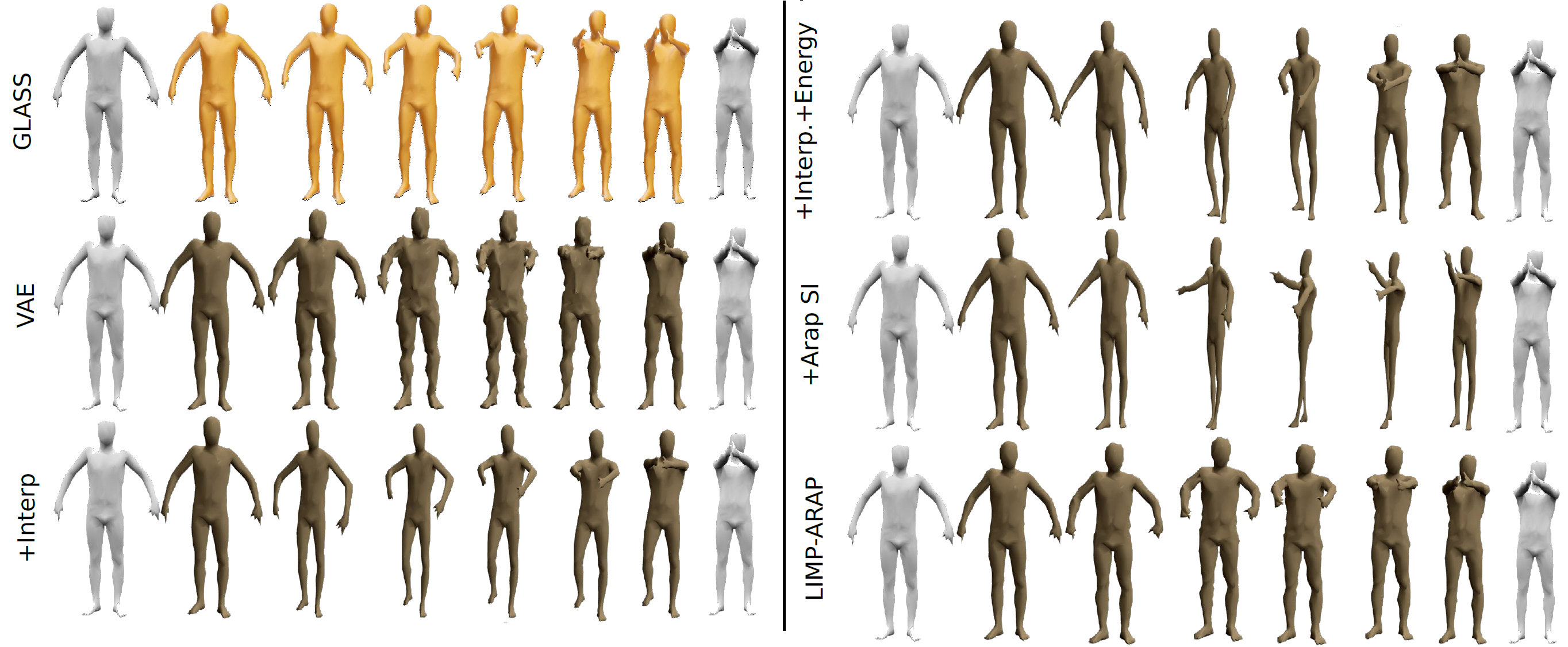} 
    \caption{We compare the interpolation results between our method, several ablations of our method, and prior work.}
    \label{fig:interpolation_comparison}
    \vspace{-\baselineskip}
\end{figure}

\paragraph{Evaluation metrics.}
We use three metrics - \textit{Coverage}, \textit{Mesh smoothness} and \textit{Interpolaiton smoothness} - to evaluate our performance. For each metric, lower values are better. The metrics are detailed in the supplemental.

\begin{figure}[t!]
    \centering
    \includegraphics[width=\linewidth,height=0.3\textheight,keepaspectratio]{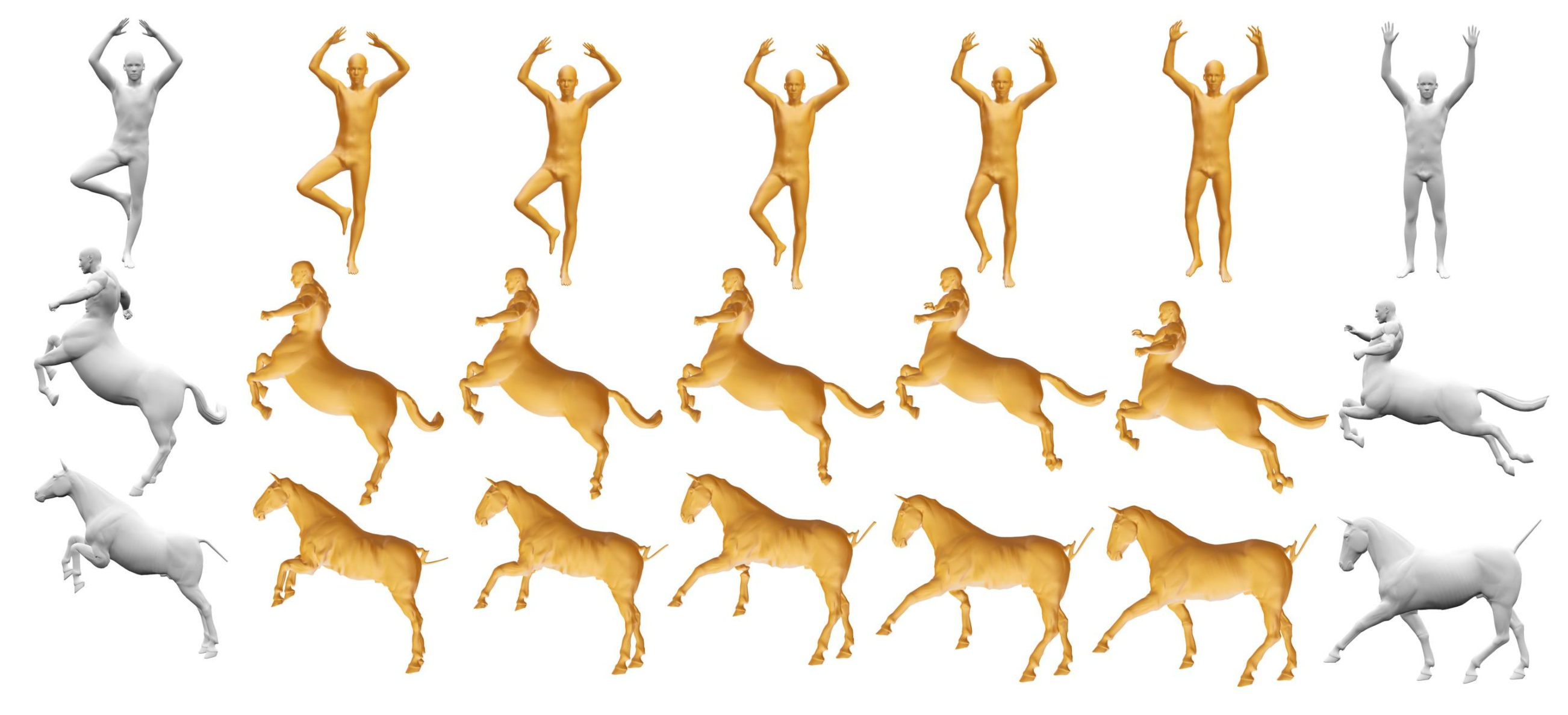}
    \caption{\emph{Interpolation results.} In gray, we show two landmark shapes. In gold, we show the decoded meshes after we linearly interpolate the latent space between these two landmarks. All models are trained on only 5 landmarks.}
    \label{fig:interpolation_our}
    \vspace{-\baselineskip}
\end{figure}

\begin{table*}[t!]
\small
\centering
  \caption{Surface smoothness and coverage with respect to excluded set, of generated samples. Lower is better. }
  \label{tab:generation}
  \begin{tabular*}{\textwidth}{r @{\extracolsep{\fill}} ccccccc}
      Data & ~Vanilla VAE~ & ~ +Interpolation ~ & ~ +Interp. +Energy ~ & LIMP & Deep-MCMC & +ARAP SI \cite{alexa2000arapinterp} & \name\\
      \hline
      \hline
      Faust-3 & 1.0 / 1.0 & 0.73 / 0.96 & 0.75 / 0.89 & 1.06 / 1.01 & 1.04 / 0.95 & 0.77 / 1.09 & \textbf{ 0.63 / 0.77}  \\
      Faust-5 & 1.0 / 1.0 & 0.65 / 1.04 & 0.62 / 0.88 & 1.0 / 0.96 & 1.06 / 0.94 & 0.62 / 0.97 & \textbf{0.59 / 0.78} \\
      Faust-7 & 1.0 / 1.0 & 1.00 / 1.51 & 0.57 / 1.61 & 1.07 / 0.93 & 1.01 / 0.96 & 0.85 / 1.84 & \textbf{0.54 / 0.81} \\
      \hline
      Centaurs-3 & 1.0 / 1.0 & 0.83 / 0.96 & 0.85 / 0.94 & 1.03 / 0.97 & 1.04 / 0.98 & 0.85 / 0.99 & \textbf{0.69 / 0.84} \\
      Centaurs-4 & 1.0 / 1.0 & 0.81 / 0.97 & 0.84 / 0.95 & 1.01 / 0.99 & 1.08 / 0.96 & 0.81 / 1.0 & \textbf{0.69 / 0.84} \\
      \hline
      Horses-3 & 1.0 / 1.0 & 0.66 / 0.95 & 0.73 / 0.91 & 1.06 / 1.03 & 1.02 / 1.05 & 0.65 / 0.94 & \textbf{0.61 / 0.89} \\
      Horses-4 & 1.0 / 1.0 & 0.65 / 0.94 & 0.68 / 0.95 & 1.03 / 0.98 & 1.04 / 1.04 & 0.62 / 1.0 & \textbf{0.60 / 0.80}  \\
      \hline
    \end{tabular*}
\end{table*}

\begin{table*}[t!]
\small
\centering
  \caption{L2 Error wrt excluded DFaust frames and reconstruction error of those excluded frames. Lower is better. }
  \label{tab:dfaust}
  \begin{tabular*}{\textwidth}{r @{\extracolsep{\fill}} ccccccc}
      Data & ~Vanilla VAE~ & ~ +Interpolation ~ & ~ +Interp. +Energy ~ & LIMP &  +ARAP SI \cite{alexa2000arapinterp} & \name\\
      \hline
      \hline
      DFaust-1 & 1.0 / 1.0 & 0.48 / 0.51 & 0.48 / 0.40 & 0.53 / 0.56 & 0.94 / 0.92 & \textbf{0.45 / 0.40} \\
      DFaust-2 & 1.0 / 1.0 & 0.56 / 0.48 & 0.51 / 0.42 & 1.06 / 1.07 & 0.84 / 0.81 & \textbf{0.47 / 0.40} \\
      DFaust-3 & 1.0 / 1.0 & 0.61 / 0.57 & 0.51 / 0.44 & 1.08 / 2.0 & 0.77 / 0.8 & \textbf{0.45 / 0.41} \\
      DFaust-4 & 1.0 / 1.0 & 0.59 / 0.35 & 0.57 / 0.5 & 0.84 / 1.73 & 0.87 / 1.14 & \textbf{0.46 / 0.32} \\
      DFaust-5 & 1.0 / 1.0 & 0.27 / 0.38 & 0.3 / 0.27 & 0.53 / 1.28 & 0.67 / 1.2 & \textbf{0.22 / 0.24} \\
    \end{tabular*}
\end{table*}

\paragraph{Method Comparison}
While existing methods are not designed to learn generative latent spaces from sparse data, we adapt them as baselines. Since ARAP projection to high resolution is specific to our method (see \ref{ss:implementation}), we compare methods using the low-res version that we train with. 

(i)~\textit{Vanilla VAE:} We train a VAE using only the sparse set of shapes, with no data augmentation.
(ii)~\textit{+Interpolation:} We generate new shapes by interpolating between all pairs of available landmarks by simply averaging coordinates of corresponding vertices.
    We interpolate such that the amount of augmented data is equivalent to what is generated with \name (2500 shapes). Then we train a VAE with these poses.
(iii)~\textit{ +Interp. +Energy:} This is an extension of the previous method.  Since raw interpolation can deviate from the underlying shape space, in addition to interpolation, we perform projection by minimizing the sum of ARAP energies with respect to both shapes in the pair.
(iv)~\textit{ LIMP-ARAP~\cite{cosmo2020limp}:} This method is motivated by the training strategy proposed in LIMP~\cite{cosmo2020limp}. They train a VAE with pairs of shapes in every iteration - for each pair, they pick a random latent code on the line between the two, decode it to a new shape, and minimize its energy. Since we only want to compare augmentation strategies we adapt LIMP to use ARAP energy.
(v)~\textit{+ ARAP Shape Interpolation \cite{alexa2000arapinterp}:} This method morphs a shape from a source pose to a target pose. The morph is rigid in the sense that local volumes are least-distorting as they vary from their source to target configurations.
(vi)~\textit{Deep-MCMC~\cite{shahbaba2019deepmcmc}:} This  method explores parameter variations via a latent space and generate samples by performing HMC steps in the latent space and decoding the generated codes. 

\begin{figure}[t!]
    \centering
    \includegraphics[width=\columnwidth]{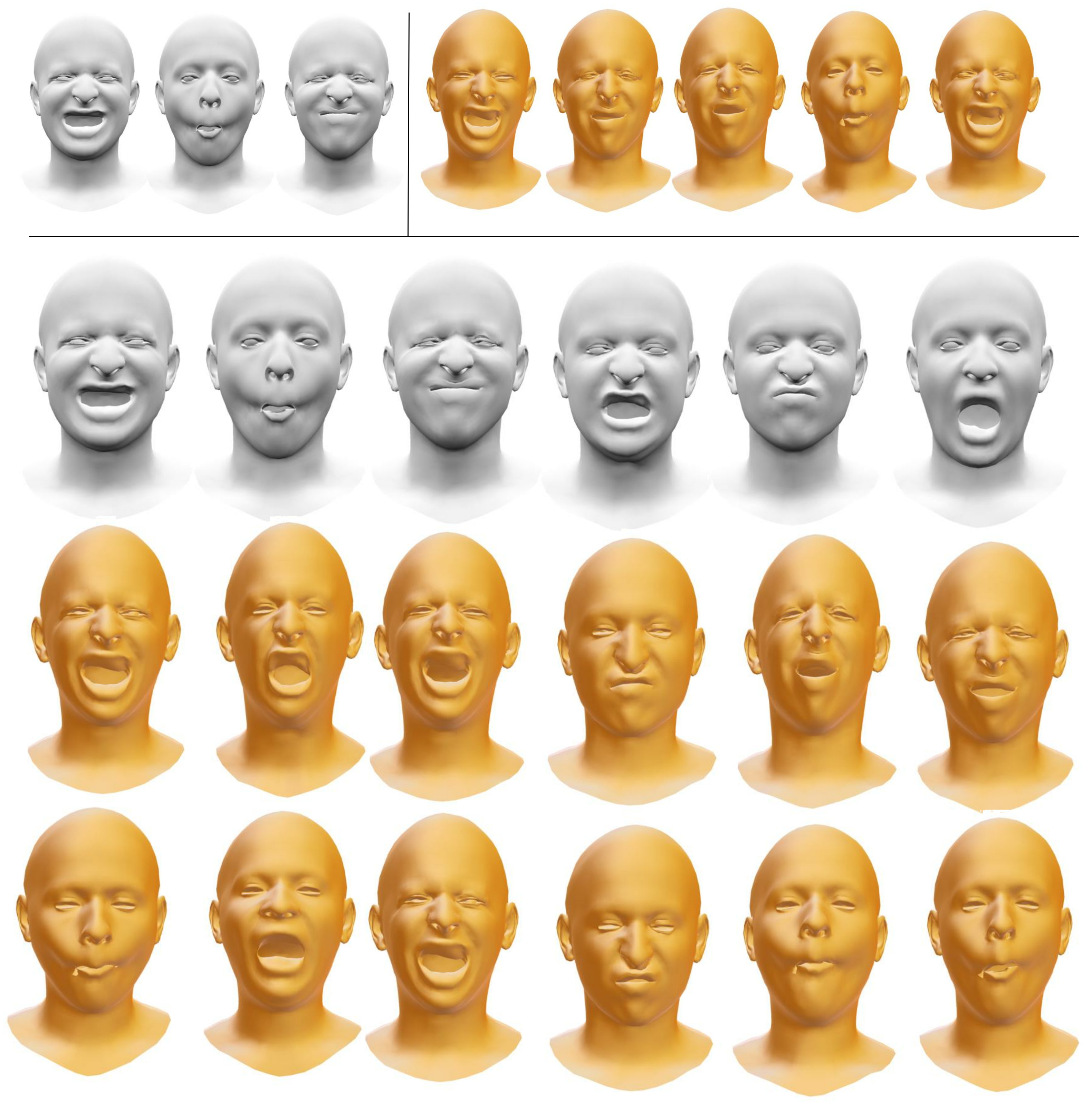} 
    \caption{In gray, are 2 subsets of 3 and 6 facial expressions from the COMA dataset. Training \name on each of them, correspondingly generates the novel expressions in gold.}
    \label{fig:faces}
    \vspace{-\baselineskip}
\end{figure}
\paragraph{Generation Experiments}

After training, we sample latent codes from a Unit Gaussian in $\mathbb{R}^K$, and decode with our decoder to generate samples (see Figures~\ref{fig:teaser}, \ref{fig:results_various} and \ref{fig:faces}). Our generated poses look substantially different from the training data and combine features from multiple input examples. 

We compare our approach to all the baselines. We sample from the unit Gaussian for all VAE-based techniques, where the only exception is Deep-MCMC where we use latent-space HMC as proposed in their work. We show qualitative results in Figure~\ref{fig:generation_comparison} and quantitative evaluations in Table~\ref{tab:generation}. Each cell reports smoothness and coverage errors, normalized based on the corresponding errors for Vanilla-VAE. Note that our method outperforms all baselines in its ability to generate novel and plausible poses (i.e., the poses from the hold-out set of the true poses).

\begin{table*}[t!]
\small
\centering
  \caption{Surface smoothness, ARAP energy, and standard deviation of inter-frame spacing between landmarks by interpolation across different datasets. All results are normalized such that Vanilla VAE is 1.0, and lower numbers are better.}
  \label{tab:interpolation}
  \begin{tabular*}{\textwidth}{r @{\extracolsep{\fill}} cccccc}
      ~Data~ & ~Vanilla VAE~ & ~ +Interpolation ~ & ~ +Interp. +Energy ~ &  ~LIMP~ & ~ +ARAP SI ~ & \name\\
      \hline
      \hline
      Faust-3 & 1.0 / 1.0 / 1.0 & 0.47 / 0.22 / 0.43 & 0.44 / 0.24 / 0.29 & 0.45 / 0.32 / 0.7 & 0.95 / 1.49 / 0.91 & \textbf{0.42}/\textbf{0.20} / \textbf{0.21} \\
Faust-5 & 1.0 / 1.0 / 1.0 & 0.53 / 0.25 / 0.49 & 0.53 / 0.25 / 0.41 & 0.52 / 0.27 / 0.5 & 0.99 / 1.13 / 0.87 & \textbf{0.51} / \textbf{0.23} / \textbf{0.38}   \\
Faust-7 & 1.0 / 1.0 / 1.0 & 0.64 / 0.31 / 0.49 & \textbf{0.57} / \textbf{0.26} / 0.62 & 0.62 / 0.32 / 0.72 & 0.84 / 0.57 / 0.86 & \textbf{0.57} / 0.3 / \textbf{0.23}  \\
Faust-10 & 1.0 / 1.0 / 1.0 & 0.71 / 0.37 / 0.3 & 0.67 / 0.43 / 0.27 & 0.66 / 0.36 / 1.52 & 0.78 / 0.48 / 0.55 & \textbf{0.55} / \textbf{0.27} / \textbf{0.54} \\
\hline
Centaurs-3 & 1.0 / 1.0 / 1.0 & 0.53 / 0.31 / 0.22 & 0.51 / 0.31 / 0.48 & 0.52 / 0.34 / 0.24 & 0.89 / 1.28 / 2.12 & \textbf{0.5} / \textbf{0.28} / \textbf{0.18} \\
Centaurs-4 & 1.0 / 1.0 / 1.0 & 0.53 / 0.24 / 0.27 & 0.52 / 0.26 / 0.17 & 0.51 / \textbf{0.25} / 0.48 & 0.88 / 0.88 / 0.8 & \textbf{0.49} / \textbf{0.25} / \textbf{0.1} \\
Centaurs-6 & 1.0 / 1.0 / 1.0 & 0.59 / 0.26 / \textbf{0.33} & 0.65 / 0.38 / 0.5 & 0.6 / 0.27 / 0.47 & 0.98 / 1.02 / 0.44 & \textbf{0.58} / \textbf{0.22} / 0.43 \\
\hline
Horses-3 & 1.0 / 1.0 / 1.0 & 0.44 / 0.39 / 0.33 & 0.46 / 0.33 / \textbf{0.28} & 0.44 / 0.4 / 0.43 & 0.9 / 1.41 / 1.28 & \textbf{0.42} / \textbf{0.24} / 0.43 \\
Horses-4 & 1.0 / 1.0 / 1.0 & 0.48 / 0.29 / 0.28 & 0.47 / 0.3 / 0.45 & \textbf{0.45} / 0.36 / 0.85 & 0.94 / 1.25 / 1.42 & 0.48 / \textbf{0.22} / \textbf{0.44} \\
Horses-8 & 1.0 / 1.0 / 1.0 & 0.55 / 0.31 / 0.55 & 0.56 / 0.38 / 0.7 & \textbf{0.53} / 0.43 / 1.29 & 0.83 / 0.76 / 0.74 &  \textbf{0.53} / \textbf{0.25} / \textbf{0.28}  \\
    \end{tabular*}
\end{table*}

\begin{table*}[t!]
\centering

  \caption{Ablation study results.}
  \small
  \begin{tabular}{r| @{\extracolsep{\fill}} || c | c || c | c || c | c }
      Data & 1: Vanilla VAE & 2: (1)+$L_\text{deform}$ & 3: (1) + perturb & 4: (2) + perturb & 5: (3)+project & 6: (4)+project \\
      \hline
      \hline
      Faust-10 & 1.0 / 1.0 / 1.0 & 0.63 / 0.39 / 0.6 & 0.61 / 0.35 / 0.6 & 0.59 / 0.32 / 0.59 & 0.56 / 0.32 / 0.57 & \textbf{0.55} / \textbf{0.27} / \textbf{0.54}  \\
      \hline
      Centaurs-6 & 1.0 / 1.0 / 1.0 & 0.69 / 0.34 / 0.54 & 0.62 / 0.31 / 0.47 & 0.59 / 0.29 / 0.47 & 0.59 / 0.27 / 0.46 & \textbf{0.58} / \textbf{0.22} / \textbf{0.43}   \\
      \hline
      Horses-8 & 1.0 / 1.0 / 1.0 & 0.66 / 0.37 / 0.43 & 0.59 / 0.37 / 0.33 & 0.56 / 0.29 / 0.32 & 0.54 / 0.27 / 0.29 & \textbf{0.53} / \textbf{0.25} / \textbf{0.28} \\
    \end{tabular}
  \label{tab:ablations}
\end{table*}

\vspace{2mm}

\paragraph{Interpolation Experiments}
We compare our method and the first five baselines by evaluating the quality of interpolations produced between all pairs of landmark shapes (we omit Deep-MCMC since it is not suitable for interpolation). We show our results in Figure~\ref{fig:interpolation_our} and comparisons in Figure~\ref{fig:interpolation_comparison} with corresponding stats in Table~\ref{tab:interpolation}. Each cell reports smoothness, ARAP score, and interpolation quality, and to make results more readable we normalize the scores using the corresponding value for Vanilla VAE.

Ours performs the best with respect to ARAP score and also yields consistently smoother shapes with fewer artifacts, both in terms of individual surfaces, as well as discontinuities in interpolation sequences. \textit{ARAP Shape Interpolation} is performing consistently worse than all baselines except Vanilla-VAE. Interpolation-based baselines sometimes yield smoother interpolations, something we would expect to be true for simpler, linear motions. However, as seen in Figure~\ref{fig:interpolation_comparison}, an outlier global rotation can disturb the otherwise smooth interpolation, which does not happen with \name. Additionally, we demonstrate that they are limited in their ability to synthesize novel plausible poses.
\begin{table}[h!]
\small
\centering
  \caption{Correspondence error on the Faust INTRA benchmark, by \name-augmenting 3D-CODED with deformations sampled from our method.}
  \begin{tabular}{r | @{\extracolsep{\fill}} c|c}
      ~ Data ~ & ~ +\name augmentation ~ & ~ Error (cm)~ \\
      \hline
      \hline
      Faust-3 & ~ 0 ~ & 26.90 \\
      Faust-3 & 3,065  & \textbf{13.18} \\
      \hline
      Faust-7 & ~ 0 ~ & 22.10 \\
      Faust-7 & 3,573 & \textbf{11.78} \\
      \hline
      SMPL-280 & ~ 0 ~ & 14.59 \\
      SMPL-280 & 40,000 & \textbf{6.85} \\
    \end{tabular}
    \label{table:3dcoded}
    \vspace{-1em}
\end{table}

\paragraph{Dynamic Faust Interpolation}
Dynamic Faust \cite{dfaust:CVPR:2017} (DFaust) provides meshed data of captured motion sequences, from which we selected 5 sequences with the most variance in vertex positions. Each sequence contains 100-200 shapes from which we select $\approx$ 5 keypoints and train \name and the baselines on these. We then evaluate interpolation by measuring the $L2$ distance between the generated interpolations and the ground-truth frames in the sequence that were excluded from training. Additionally, we measure the reconstruction error of the excluded frames. The results are compared in Table  \ref{tab:dfaust}. Our method significantly outperforms the baselines.

\paragraph{Using \name for Learning Correspondences}

We now evaluate our data augmentation technique on the practical task of learning 3D correspondences between shapes. We pick a state-of-the-art correspondence learning method, 3D-CODED~\cite{groueix2018coded}, as a reference. Originally, this method was trained on 230k deformations, most of them sampled from the SMPL model~\cite{SMPL:2015} and 30k synthetic augmentations. We evaluate how well this method could perform with a smaller training set, with and without the proposed augmentation.

We train 3D-CODED using 280 sampled shapes from SMPL~\cite{SMPL:2015} that are augmented with a number of additional deformations generated from our model (see  Table~\ref{table:3dcoded}). Ours consistently provides a significant improvement over training 3D-CODED with the original landmarks. For reference, the correspondence error of 3D-CODED trained on the full 230k pose dataset is 1.98cm.

\paragraph{Ablation study}

In this section we evaluate the contribution of various steps in \name to our interpolation metrics. We evaluate on the Faust-10, Centaurs-6 and Horses-8 datasets. Since these samples are all that is available we do not have a hold-out set, so we do not measure coverage.  Starting with Vanilla-VAE (that only uses $L_\text{Reconstruction}$ and $L_\text{Gaussian}$), we first add the deformation energy loss ($L_\text{Deformation}$), which is ARAPReg \cite{huang2021arapreg}. Table~\ref{tab:ablations}(1,2) shows this improves all metrics.

Next, we consider our perturbation strategy (Section~\ref{sec:augmentation}~i, ii) and add it to both vanilla and energy-guided VAE (Table~\ref{tab:ablations}.3, \ref{tab:ablations}.4).  We observe that $L_\text{Deformation}$ performs better because it makes the latent-space conducive for sampling low energy shapes. Having $L_\text{Deformation}$ helps our perturbation strategy find low energy shapes that are suitable for our projection step. Due to this, we discover shapes with energy as low as $0.001$, while without it, the discovered shapes can have energy $> 0.1$. This difference helps the subsequent projection step converge faster to our required threshold of $10^{-5}$.

Finally, we look at the projection step (Section~\ref{sec:augmentation}~iii). We add it to both baseline techniques that have perturbation, and report results in Table~\ref{tab:ablations}.5, \ref{tab:ablations}.6, where column (6) corresponds to our final method. Adding the projection step improves the smoothness and ARAP scores. After projection, our shapes have very low ARAP, in the order of $10^{-5}$. Since these are added back to the training set, we observe that  perturbation steps in future iterations find lower energy shapes. This further improves convergence of the projection step in future iterations.  Overall $L_\text{Deformation}$ helps both the perturbation and projection steps converge faster to low energy shapes, and since projected shapes are encoded again by training, both perturbation and projection steps require fewer iterations.

\section{Conclusion}

\name is shown to be an effective generative technique for   3D shape deformations, relying solely on a  handful of examples and a given deformation energy. The main limitation of our method is its reliance on a given mesh with vertex correspondences, preventing its use on examples with different triangulations, and we set the goal of generalizing it to arbitrary geometries as important future work.

We believe our proposed technique opens many future directions. There are many other deformation energies that could be explored; e.g., densely sampling conformal (or quasi-conformal) deformations  from a given sparse set can be an extremely interesting followup. More broadly, replacing the deformation energy with learned energies, such as the output of an image-discriminator, may enable generating plausible images, given a very sparse set of examples.  

\paragraph{Acknowledgement} This project has received funding from the UCL AI Center, gifts from Adobe Research, and EU Marie Skłodowska-Curie grant agreement No 956585.

{\small
\bibliographystyle{ieee_fullname}
\bibliography{glass}

\begin{thebibliography}{10}\itemsep=-1pt

\bibitem{alexa2000arapinterp}
Marc Alexa, Daniel Cohen-Or, and David Levin.
\newblock As-rigid-as-possible shape interpolation.
\newblock In {\em SIGGRAPH}, 2000.

\bibitem{allen2003bodyshape}
Brett Allen, Brian Curless, and Zoran Popovi\'{c}.
\newblock The space of human body shapes: Reconstruction and parameterization
  from range scans.
\newblock In {\em SIGGRAPH}, 2003.

\bibitem{anguelov2005scape}
Dragomir Anguelov, Praveen Srinivasan, Daphne Koller, Sebastian Thrun, Jim
  Rodgers, and James Davis.
\newblock {SCAPE}: Shape completion and animation of people.
\newblock In {\em SIGGRAPH}, 2005.

\bibitem{blanz1999morph3d}
Volker Blanz and Thomas Vetter.
\newblock A morphable model for the synthesis of {3D} faces.
\newblock In {\em SIGGRAPH}, 1999.

\bibitem{bogo2014faust}
Federica Bogo, Javier Romero, Matthew Loper, and Michael~J. Black.
\newblock {FAUST}: Dataset and evaluation for {3D} mesh registration.
\newblock In {\em CVPR}, 2014.

\bibitem{dfaust:CVPR:2017}
Federica Bogo, Javier Romero, Gerard Pons-Moll, and Michael~J. Black.
\newblock Dynamic {FAUST}: {R}egistering human bodies in motion.
\newblock In {\em CVPR}, 2017.

\bibitem{botsch2008linear}
Mario Botsch and Olga Sorkine.
\newblock On linear variational surface deformation methods.
\newblock {\em TVCG}, 14(1), 2008.

\bibitem{bronstein2008numerical}
Alexander~M Bronstein, Michael~M Bronstein, and Ron Kimmel.
\newblock {\em Numerical geometry of non-rigid shapes}.
\newblock Springer Science \& Business Media, 2008.

\bibitem{MMR_1998}
Jaime Carbonell and Jade Goldstein.
\newblock The use of {MMR}, diversity-based reranking for reordering documents
  and producing summaries.
\newblock In {\em SIGIR}, 1998.

\bibitem{chaudhary2020nlpaug}
Amit Chaudhary.
\newblock A visual survey of data augmentation in {NLP}, 2020.
\newblock \url{https://amitness.com/2020/05/data-augmentation-for-nlp}.

\bibitem{chaudhuri2020structgen}
Siddhartha Chaudhuri, Daniel Ritchie, Jiajun Wu, Kai Xu, and Hao Zhang.
\newblock Learning generative models of {3D} structures.
\newblock {\em Comput. Graph. For. (Eurographics STAR)}, 2020.

\bibitem{cosmo2020limp}
Luca Cosmo, Antonio Norelli, Oshri Halimi, Ron Kimmel, and Emanuele Rodol{\`a}.
\newblock {LIMP}: Learning latent shape representations with metric
  preservation priors.
\newblock {\em ECCV}, 2020.

\bibitem{gadelha2020handles}
Matheus Gadelha, Giorgio Gori, Duygu Ceylan, Radomir M{\v{e}}ch, Nathan Carr,
  Tamy Boubekeur, Rui Wang, and Subhransu Maji.
\newblock Learning generative models of shape handles.
\newblock In {\em CVPR}, 2020.

\bibitem{gaussian_reg}
Matheus Gadelha, Rui Wang, and Subhransu Maji.
\newblock Multiresolution tree networks for {3D} point cloud processing.
\newblock {\em CoRR}, abs/1807.03520, 2018.

\bibitem{gao2021acap}
Lin Gao, Yu-Kun Lai, Jie Yang, Ling-Xiao Zhang, Shihong Xia, and Leif Kobbelt.
\newblock Sparse data driven mesh deformation.
\newblock {\em TVCG}, 27(3), 2021.

\bibitem{gao2019sdmnet}
Lin Gao, Jie Yang, Tong Wu, Yu-Jie Yuan, Hongbo Fu, Yu-Kun Lai, and Hao Zhang.
\newblock {SDM-NET}: Deep generative network for structured deformable mesh.
\newblock {\em ACM Trans. Graph.}, 38(6), 2019.

\bibitem{goodfellow2015adversarial}
Ian Goodfellow, Jonathon Shlens, and Christian Szegedy.
\newblock Explaining and harnessing adversarial examples.
\newblock In {\em ICLR}, 2015.

\bibitem{groueix2018coded}
Thibault Groueix, Matthew Fisher, Vladimir~G. Kim, Bryan~C. Russell, and
  Mathieu Aubry.
\newblock {3D-CODED}: {3D} correspondences by deep deformation.
\newblock {\em ECCV}, 2018.

\bibitem{helein2008harmonic}
Fr\'ed\'eric H\'elein and John~C. Wood.
\newblock {\em Handbook of Global Analysis}, chapter Harmonic Maps.
\newblock 2008.

\bibitem{huang2021arapreg}
Qixing Huang, Xiangru Huang, Bo Sun, Zaiwei Zhang, Junfeng Jiang, and
  Chandrajit Bajaj.
\newblock {ARAPReg}: An as-rigid-as possible regularization loss for learning
  deformable shape generators.
\newblock {\em CoRR}, abs/2108.09432, 2021.

\bibitem{huang2009modal}
Qi-Xing Huang, Martin Wicke, Bart Adams, and Leonidas Guibas.
\newblock Shape decomposition using modal analysis.
\newblock {\em Computer Graphics Forum}, 28(2), 2009.

\bibitem{jacobson2011bbw}
Alec Jacobson, Ilya Baran, Jovan Popovi\'{c}, and Olga Sorkine.
\newblock Bounded biharmonic weights for real-time deformation.
\newblock {\em ACM Trans. Graph.}, 30(4), 2011.

\bibitem{jacobson2014skinning}
Alec Jacobson, Zhigang Deng, Ladislav Kavan, and J.~P. Lewis.
\newblock Skinning: Real-time shape deformation.
\newblock In {\em SIGGRAPH Courses}, 2014.

\bibitem{ju2005meanvalue}
Tao Ju, Scott Schaefer, and Joe Warren.
\newblock Mean value coordinates for closed triangular meshes.
\newblock In {\em SIGGRAPH}, 2005.

\bibitem{kilian2007shape}
Martin Kilian, Niloy~J. Mitra, and Helmut Pottmann.
\newblock Geometric modeling in shape space.
\newblock {\em ACM Trans. Graph.}, 26(3), 2007.

\bibitem{kingma2014autoencoding}
Diederik~P. Kingma and Max Welling.
\newblock Auto-encoding variational bayes.
\newblock In {\em ICLR}, 2014.

\bibitem{laga2018nonrigid}
Hamid Laga.
\newblock A survey on non-rigid {3D} shape analysis.
\newblock {\em CoRR}, abs/1812.10111, 2018.

\bibitem{levy2002lscm}
Bruno L\'evy, Sylvain Petitjean, Nicolas Ray, and J\'er\^{o}me Maillot.
\newblock Least squares conformal maps for automatic texture atlas generation.
\newblock In {\em SIGGRAPH}, 2002.

\bibitem{lewis2014blendshape}
J.~P. Lewis, Ken Anjyo, Taehyun Rhee, Mengjie Zhang, Fred Pighin, and Zhigang
  Deng.
\newblock Practice and theory of blendshape facial models.
\newblock {\em Eurographics State of the Art Reports}, 2014.

\bibitem{SMPL:2015}
Matthew Loper, Naureen Mahmood, Javier Romero, Gerard Pons-Moll, and Michael~J.
  Black.
\newblock {SMPL}: A skinned multi-person linear model.
\newblock {\em ACM Trans. Graphics (Proc. SIGGRAPH Asia)}, 34(6), 2015.

\bibitem{ranjan2018meshae}
Anurag Ranjan, Timo Bolkart, Soubhik Sanyal, and Michael~J Black.
\newblock Generating {3D} faces using convolutional mesh autoencoders.
\newblock In {\em ECCV}, 2018.

\bibitem{rong2008spectral}
Guodong Rong, Yan Cao, and Xiaohu Guo.
\newblock Spectral mesh deformation.
\newblock {\em The Visual Computer}, 24, 2008.

\bibitem{shahbaba2019deepmcmc}
Babak Shahbaba, Luis~Martinez Lomeli, Tian Chen, and Shiwei Lan.
\newblock {Deep Markov Chain Monte Carlo}.
\newblock {\em CoRR}, abs/1910.05692, 2019.

\bibitem{shankar2018crossgrad}
Shiv Shankar, Vihari Piratla, Soumen Chakrabarti, Siddhartha Chaudhuri, Preethi
  Jyothi, and Sunita Sarawagi.
\newblock Generalizing across domains via cross-gradient training.
\newblock In {\em ICLR}, 2018.

\bibitem{shorten2019imgaug}
Connor Shorten and Taghi~M. Khoshgoftaar.
\newblock A survey on image data augmentation for deep learning.
\newblock {\em Journal of Big Data}, 6(1), 2019.

\bibitem{sorkine2007arap}
Olga Sorkine and Marc Alexa.
\newblock As-rigid-as-possible surface modeling.
\newblock In {\em SGP}, 2007.

\bibitem{tan2018meshvae}
Qingyang Tan, Lin Gao, Yu-Kun Lai, and Shihong Xia.
\newblock Variational autoencoders for deforming {3D} mesh models.
\newblock In {\em CVPR}, 2018.

\bibitem{tan2020thinshell}
Qingyang Tan, Zherong Pan, Lin Gao, and Dinesh Manocha.
\newblock Realtime simulation of thin-shell deformable materials using
  {CNN}-based mesh embedding.
\newblock {\em IEEE Robotics and Automation Letters}, 5(2), 2020.

\bibitem{tycowicz2015interp}
Christoph Von-Tycowicz, Christian Schulz, Hans-Peter Seidel, and Klaus
  Hildebrandt.
\newblock Real-time nonlinear shape interpolation.
\newblock {\em ACM Trans. Graph.}, 34(3), 2015.

\bibitem{wang2019neural}
Yifan Wang, Noam Aigerman, Vladimir~G. Kim, Siddhartha Chaudhuri, and Olga
  Sorkine-Hornung.
\newblock Neural cages for detail-preserving {3D} deformations.
\newblock In {\em CVPR}, 2020.

\bibitem{wen2021timeaug}
Qingsong Wen, Liang Sun, Fan Yang, Xiaomin Song, Jingkun Gao, Xue Wang, and
  Huan Xu.
\newblock Time series data augmentation for deep learning: A survey.
\newblock {\em CoRR}, abs/2002.12478, 2021.

\bibitem{yuan2021revisit}
Yu-Jie Yuan, Yu-Kun Lai, Tong Wu, Lin Gao, and Ligang Liu.
\newblock A revisit of shape editing techniques: from the geometric to the
  neural viewpoint.
\newblock {\em CoRR}, abs/2103.01694, 2021.

\bibitem{yumer2014handles}
Mehmet~Ersin Yumer and Levent~Burak Kara.
\newblock Co-constrained handles for deformation in shape collections.
\newblock {\em ACM Trans. Graph.}, 33(6), 2014.

\end{thebibliography}
}

\clearpage
\begin{center}
    {\Large \bf Appendix for \name: Geometric Latent Augmentation for Shape Spaces}
\end{center}

\appendix
\section{Evaluation metrics.} 

We use the following  to evaluate performance:
\\
\noindent
\textit{(i) Coverage:} While it is difficult to evaluate whether generated poses are meaningful and diverse, we propose using the holdout data ($S_H$) that was not part of the input exemplars to see if the newly generated poses $L_G$ cover every holdout example: \\ 
$
 M_\text{coverage} := \sum_{s \in S_H} {\min_{g \in L_G}{D(s, g)}}/{|S_H|},
$
where $D(s,g)$ is the average Euclidean distance between corresponding vertices of shapes $s$ and $g$. ~\\
\noindent
\textit{(ii) Mesh smoothness:} This metric measures how well a method preserves the original intrinsic structure of the mesh. We compute the mean curvature obtained from the discrete Laplace-Beltrami operator: \\ 
$
  M_\text{smoothness} := \sum_{i=1}^{N}  \lVert \Delta(V_i) \rVert / 2
$
where $\Delta$ is the area-normalized cotangent Laplace-Beltrami operator and $N$ is the number of mesh vertices.
\\
\noindent
\textit{(iii) Interpolation smoothness:} In addition to measuring quality of individual meshes, we also evaluate the quality of interpolations between pairs of shapes. Since none of the existing methods guarantee a clear relationship between the distances in the latent space and differences between their 3D counterparts, we first densely sample 1000 poses between pairs of landmark deformations and then keep a subset of 30 poses so that they have approximately equal average Euclidean distance between subsequent frames. We then measure the standard deviation of these Euclidean distances, as a way to penalize interpolations that yield significant jumps between frames. While this measure is imperfect (e.g., variance could decrease as we increase the sampling rate), we found it to stabilize in practice after 1000 poses (we sampled up to 3000), which suggests that denser sampling would not reveal new poses in the latent space.

\section{Variable step-size $\alpha$}
\label{sec:stepsize}

Let $f$ be the energy function and $l$ be the latent code corresponding to a zero/low energy shape in the training set, and $l_t$ be the latent code obtained from the update in Equation 6 in the main paper under a small (update) step (i.e., $\|l_t - l \|$ can be considered to be infinitesimal). Assuming that the deformation energy is C2 continuous (and single valued), $f(l_t)$ can be approximated using Taylor series expansion of the up to the \nth{2} order as,
\begin{equation} \label{eq:taylor}
  f(l_t) \approx f(l) + (l_t-l)^T \nabla_l f(l_t) + \frac{1}{2} (l_t-l)^{T} H (l_t-l),    
\end{equation}
where $H$ is the Hessian from Equation 5. Note that $H$ denotes the Hessian term with respect to the current pose $l$.  Since $(l_t-l)$ and $\nabla_l f(l_t)$ are orthogonal, $(l_t-l)^T \nabla_l f(l_t) = 0$. The update step can be expressed in terms of the local eigenvectors as (see Equation 6 in the main paper), 
\[
 (l_t-l) = \alpha \sum_{i=1}^{i=k} {\hat{\beta}_i U^{\uparrow}_{i}(H)}.
\]

Further, since eigenvalues and eigenvectors are related as $ H U^{\uparrow}_{i}(H) = \lambda_i U^{\uparrow}_{i}(H) $ and the vectors $U^{\uparrow}_{i}(H)$ have unit length, we obtain, 
\[
  f(l_t) \approx f(l) + \frac{1}{2} \alpha^2 \sum_{i=1}^{i=k} \hat{\beta}_{i}^{2} \lambda_i.
\]

By setting an upper bound on the change in deformation energy $f(l_t) - f(l) \le \delta$, we obtain the relation for $\alpha$ as,
\[
\alpha \le  \sqrt{\frac{2\delta}{\sum_{i=1}^{k} \hat{\beta}_{i}^{2} \lambda_i}}.
\]

\section{Additional Generated Samples}
\begin{figure}[h!]
    \centering
    \includegraphics[width=\columnwidth]{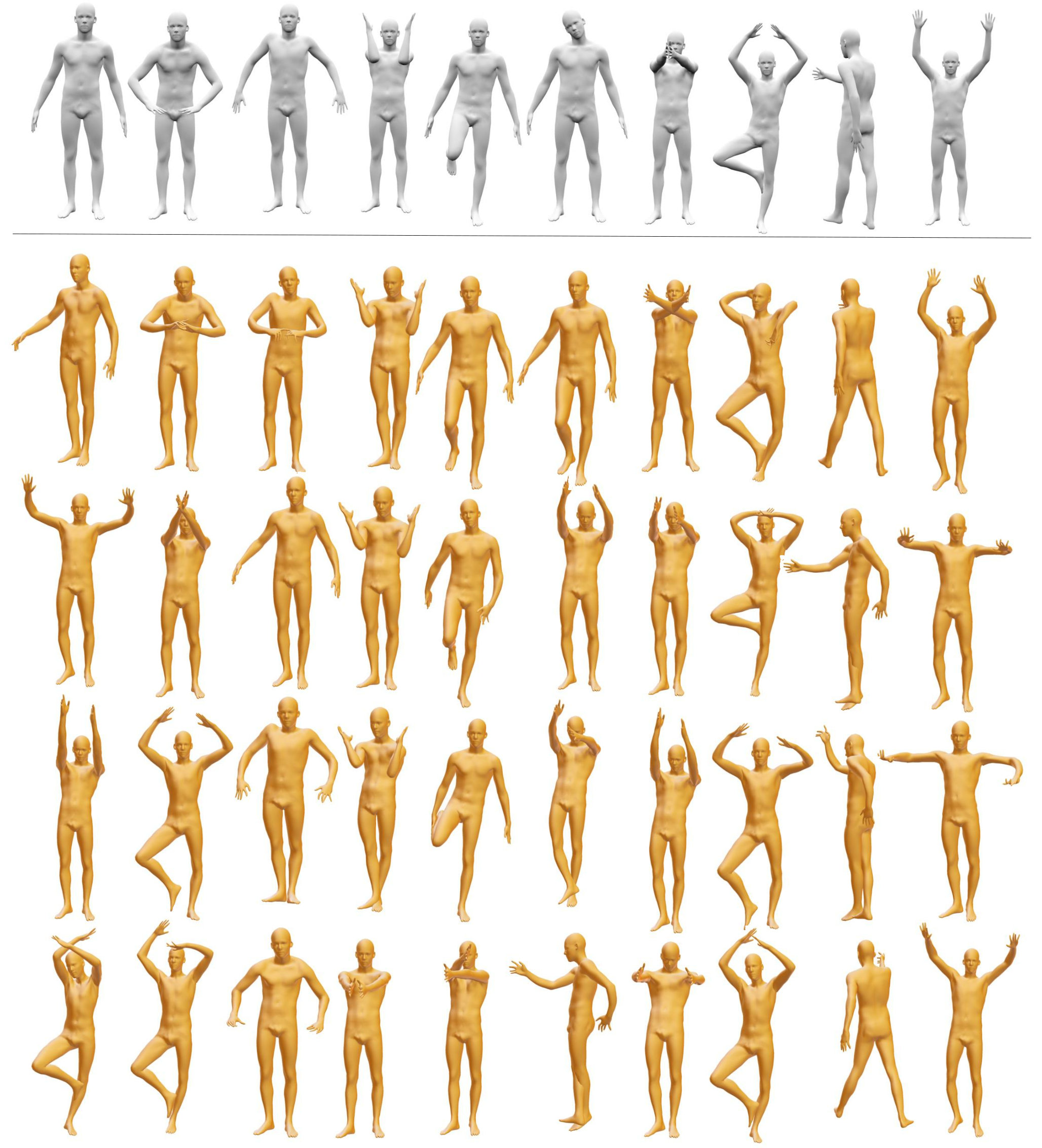}
    \caption{We show some of the samples generated by \name (gold), from only 10 Faust poses (gray). Several poses of the limbs are unseen in the training set - crossed arms, long leg strides, half-lowered arms.}
    \label{fig:faust_gen}
\end{figure}

\begin{figure}[h!]
    \centering
    \includegraphics[width=\columnwidth]{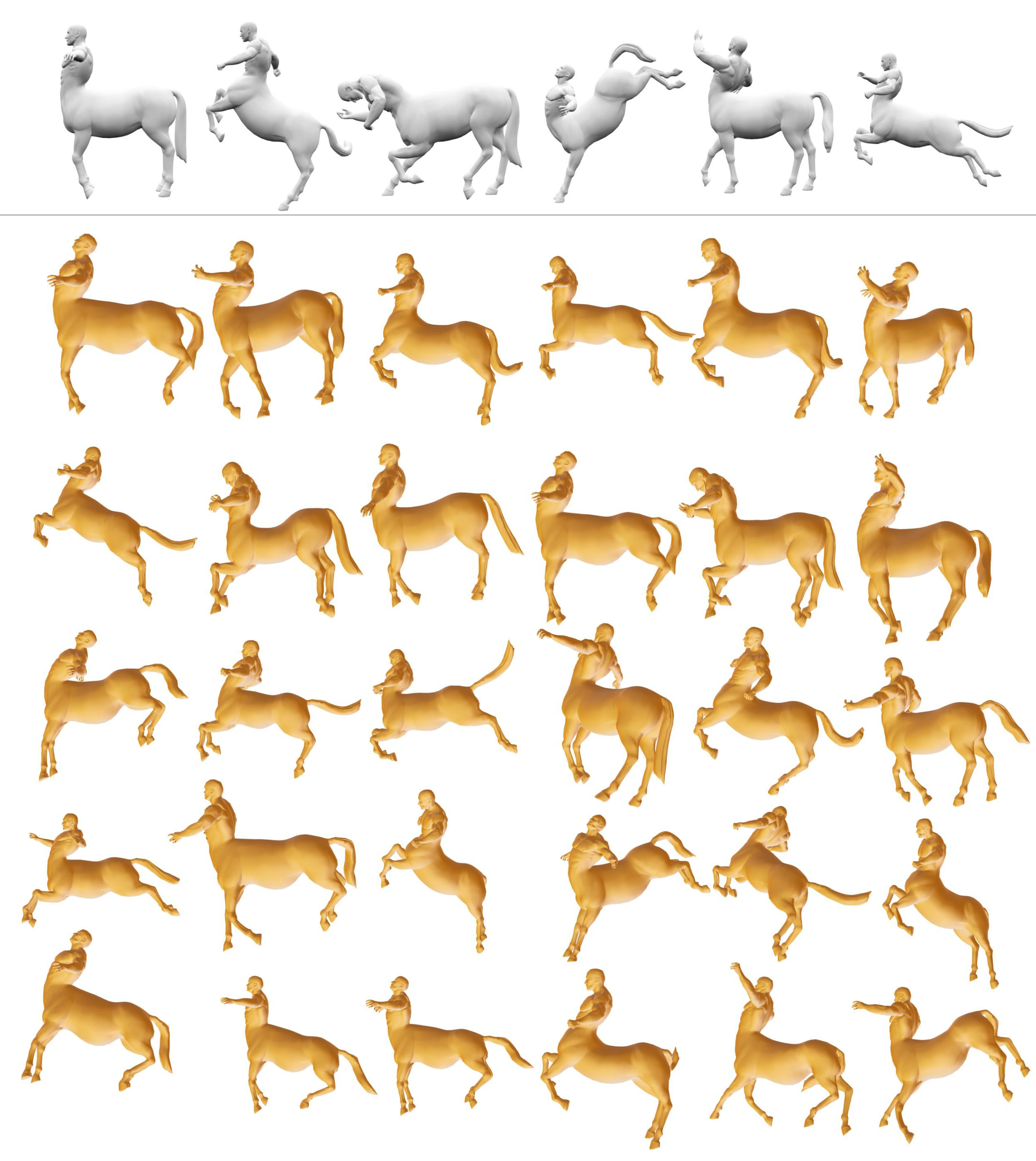}
    \caption{We show some of the samples generated by \name (gold), from only 6 Centaur poses (gray). We see novel poses like bent back legs and torso facing upwards.}
    \label{fig:centaurs_gen}
\end{figure}

\begin{figure}[h!]
    \centering
    \includegraphics[width=\columnwidth]{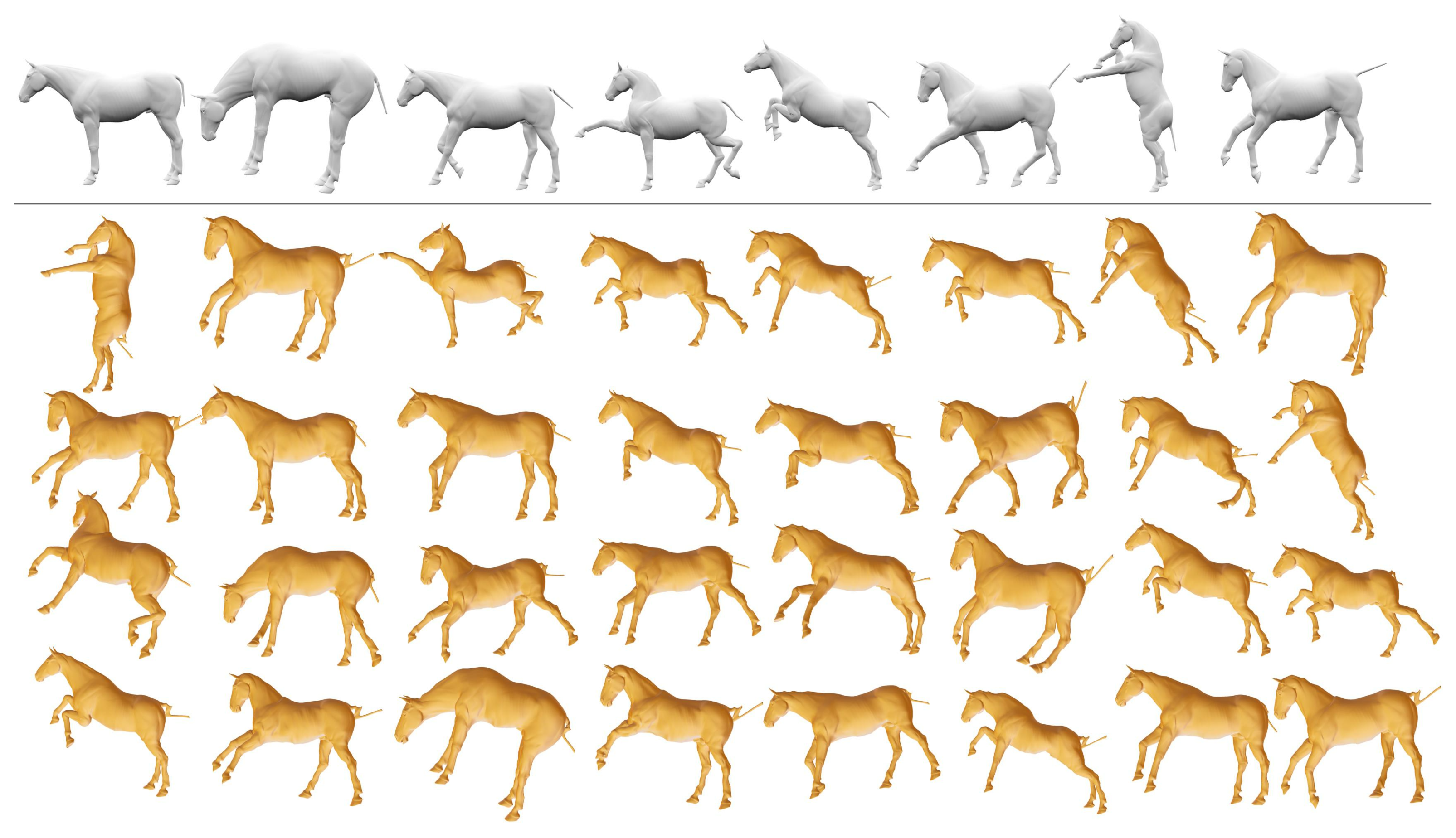}
    \caption{We show some of the samples generated by \name (gold), from only 8 Horse poses (gray). We see novel poses like front legs raised beyond what's seen in the training set, upright torso and back legs stretching farther. }
    \label{fig:horse_gen}
\end{figure}

Figures~\ref{fig:faust_gen}, \ref{fig:centaurs_gen}, and  \ref{fig:horse_gen} show some example generated deformations starting from sparse sets of Faust, Centaur, and Horse models, respectively. 

\newpage
\section{Additional Interpolation Examples}
\begin{figure}[h!]
    \centering
    \includegraphics[width=\columnwidth]{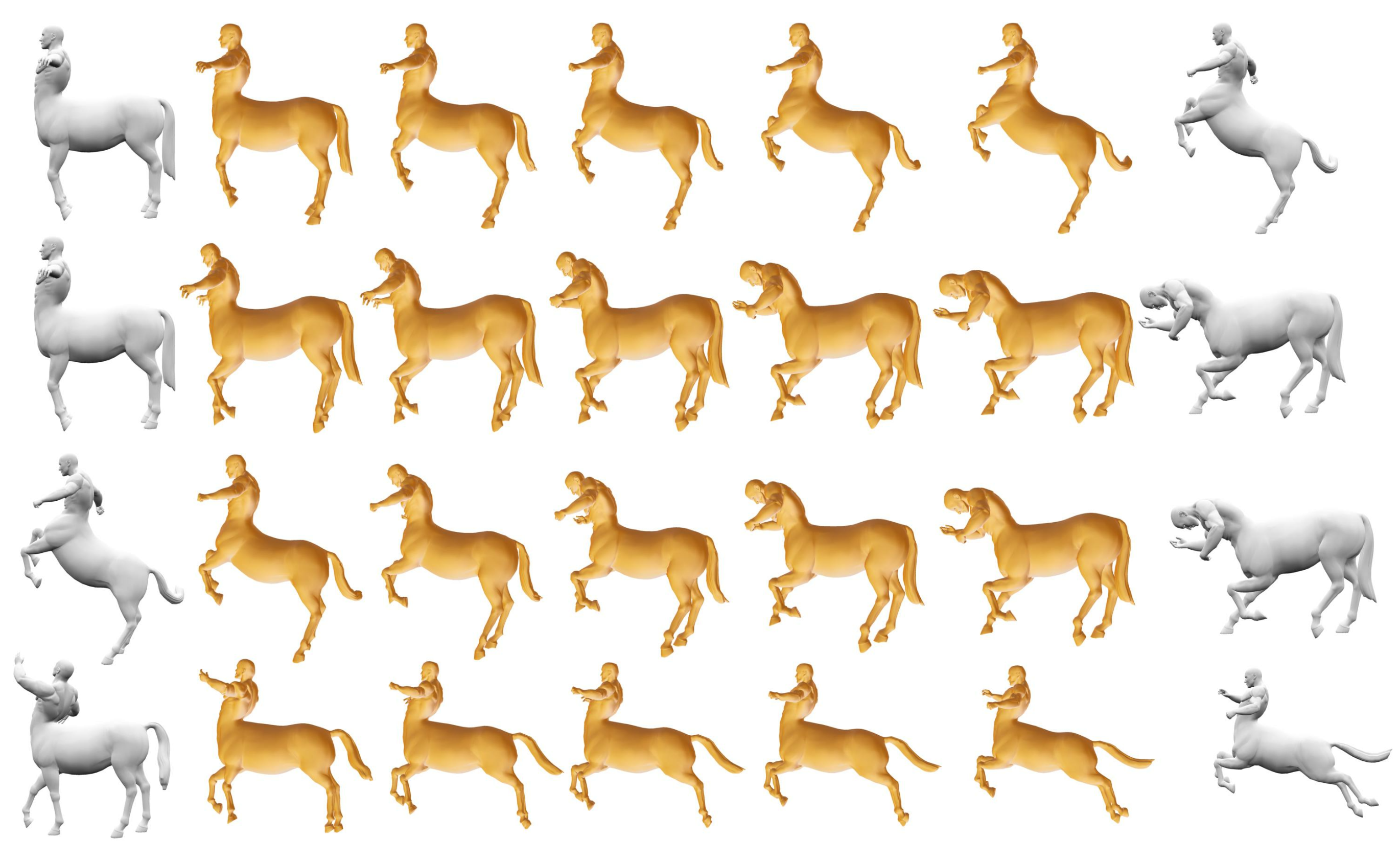}
    \caption{Interpolated shapes (gold) between 2 Centaurs Poses (gray) inside the latent space generated using \name.}
    \label{fig:centaurs_interp}
\end{figure}

\begin{figure}[h!]
    \centering
    \includegraphics[width=\columnwidth]{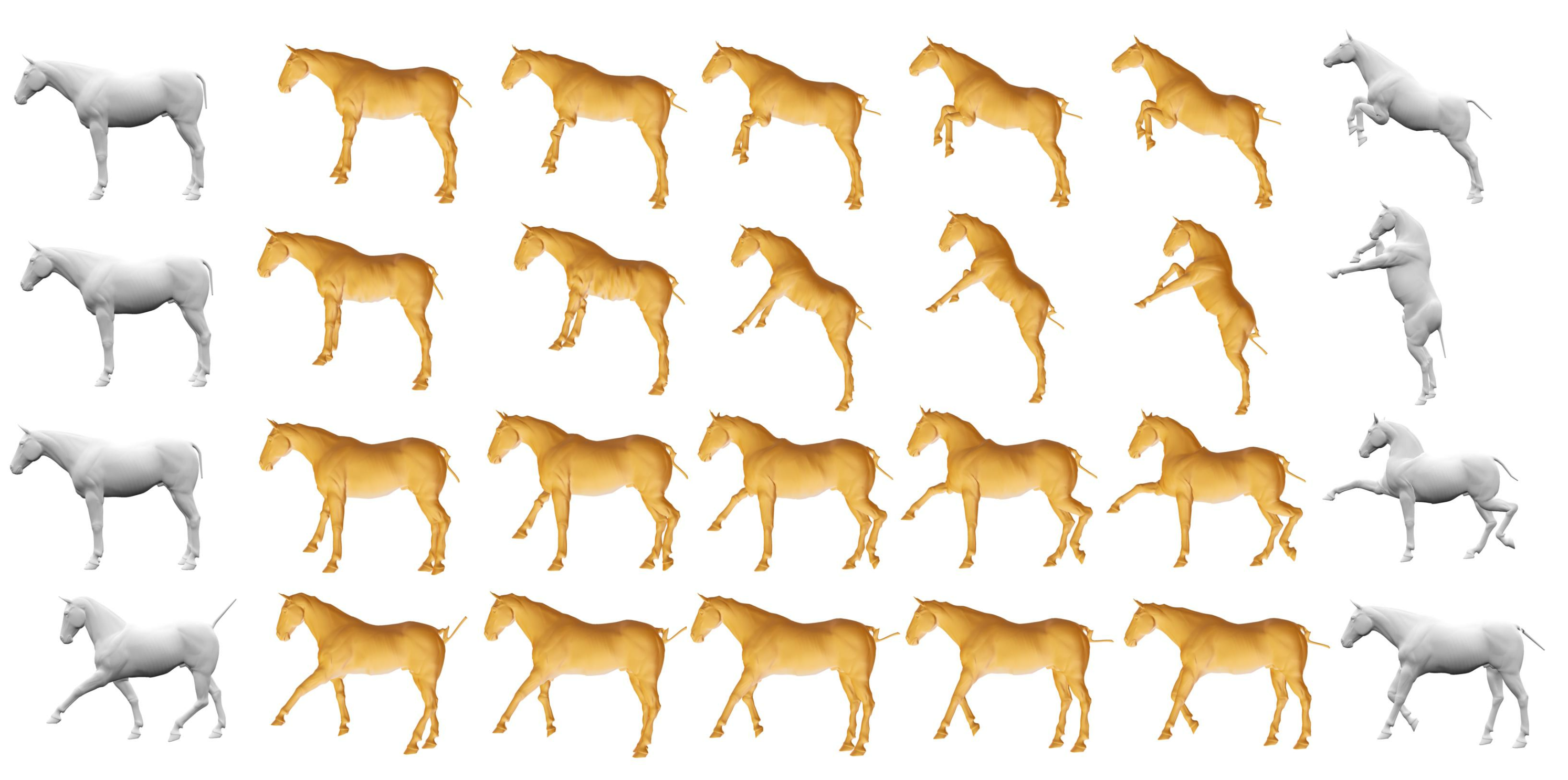}
    \caption{Interpolated shapes (gold) between 2 Horse Poses (gray) inside the latent space generated using \name.}
    \label{fig:horses_interp}
\end{figure}

\begin{figure}[h!]
    \centering
    \includegraphics[width=0.8\columnwidth]{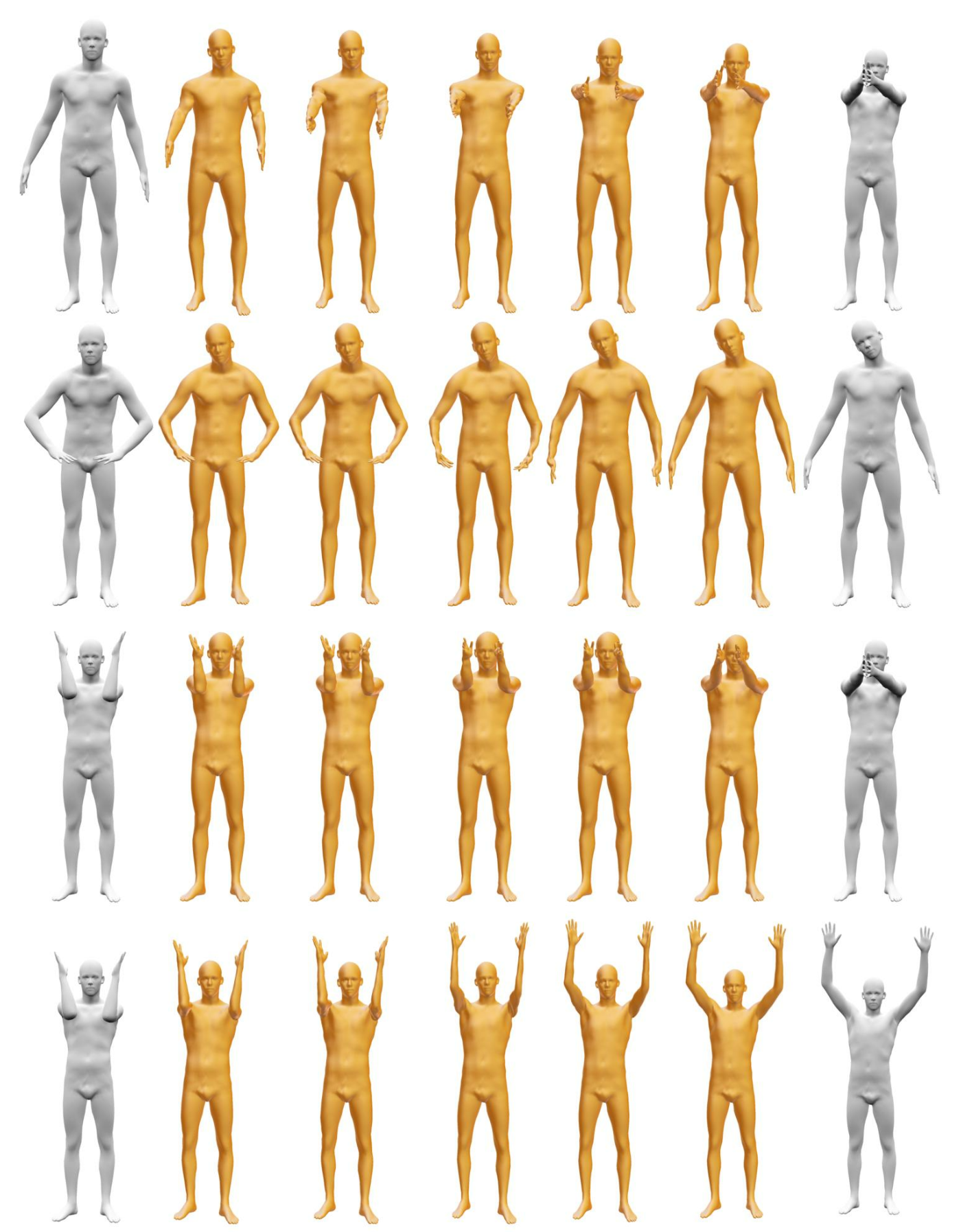}
    \caption{Interpolated shapes (gold) between 2 Faust Poses (gray) inside the latent space generated using \name.}
    \label{fig:faust_interp}
    \vspace{-\baselineskip}
\end{figure}

Figures~\ref{fig:centaurs_interp}, \ref{fig:horses_interp} and \ref{fig:faust_interp} show example interpolated shapes between end poses~(shown in gray), using the discovered latent space revealed by \name.

\section{Dataset Images}

We include images of the various input datasets we used to highlight the diversity of pose variations in the input. Note that the datasets are all very sparse, consisting of 3-10 models. 

\begin{figure}[h!]
    \centering
    \includegraphics[width=0.8\columnwidth]{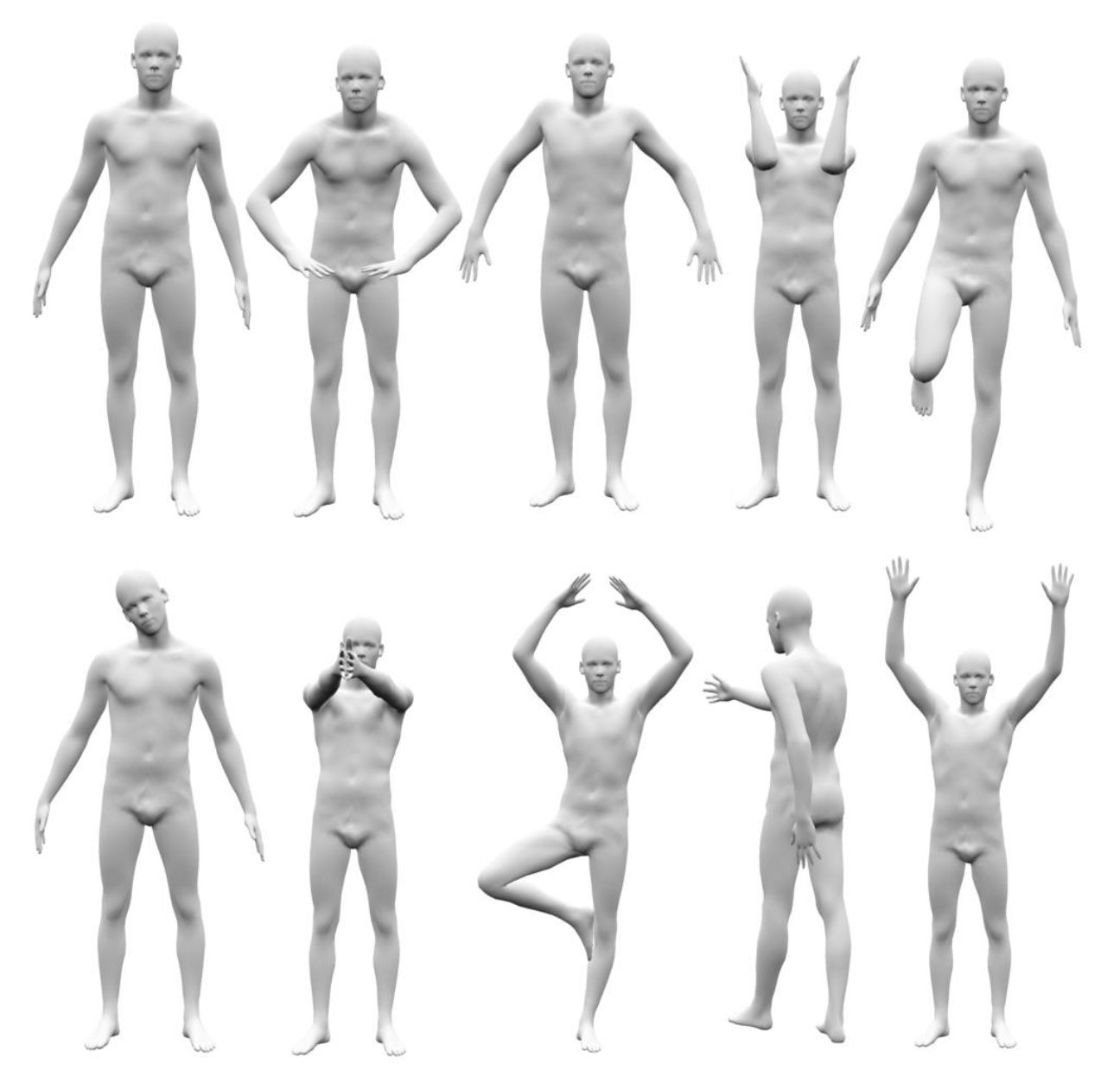}
    \caption{The Faust-10 dataset with 10 poses. Please refer to Figure~\protect\ref{fig:faust3} for the corresponding even sparser versions of the dataset used in our experiments. }
    \label{fig:faust10}
\end{figure}

\begin{figure}[h!]
    \centering
    \includegraphics[width=\columnwidth]{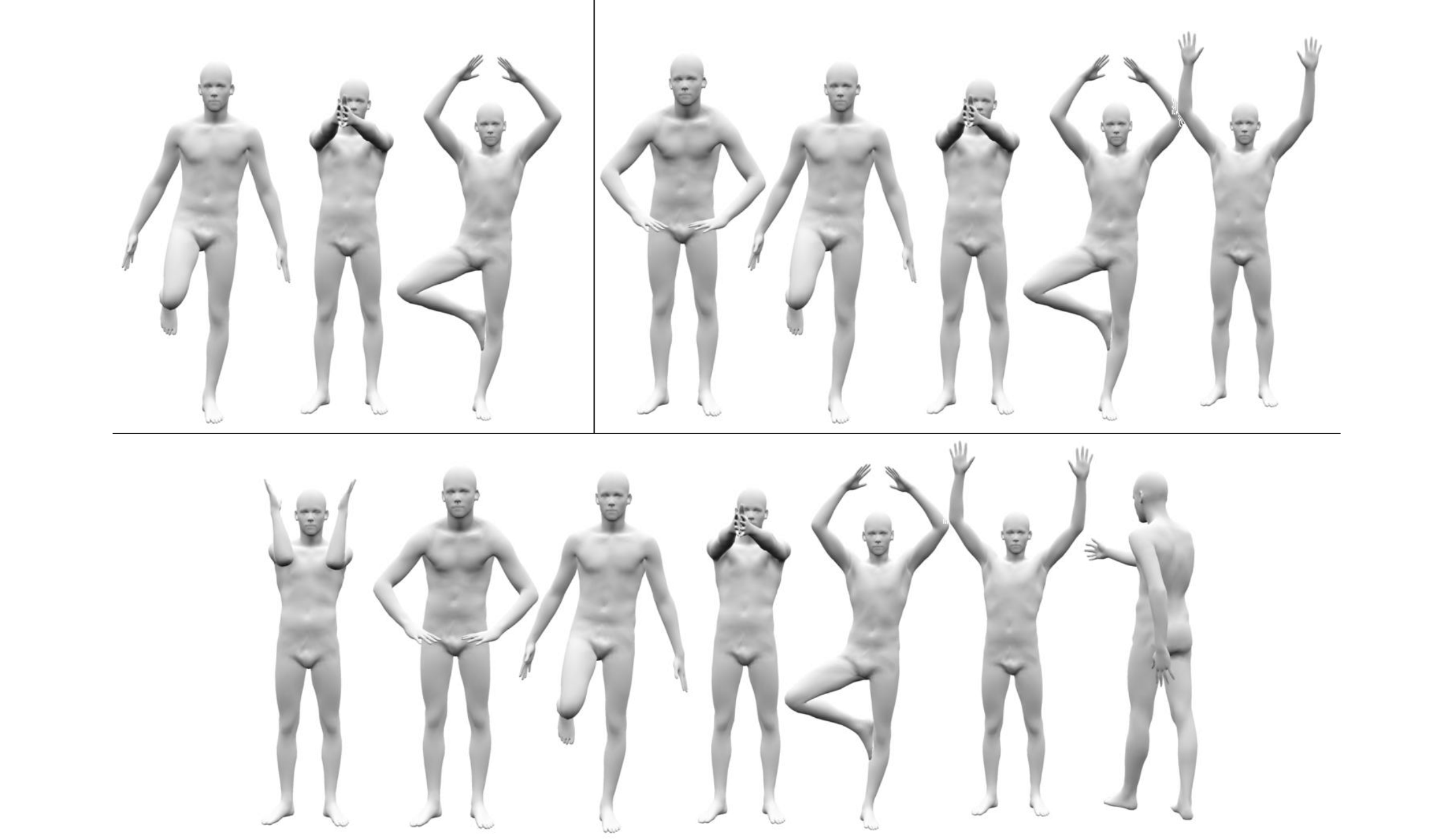}
    \caption{Faust-3 (top left), Faust-5 (top right) and Faust-7 (bottom).}
    \label{fig:faust3}
\end{figure}

\begin{figure}[h!]
    \centering
    \includegraphics[width=\columnwidth]{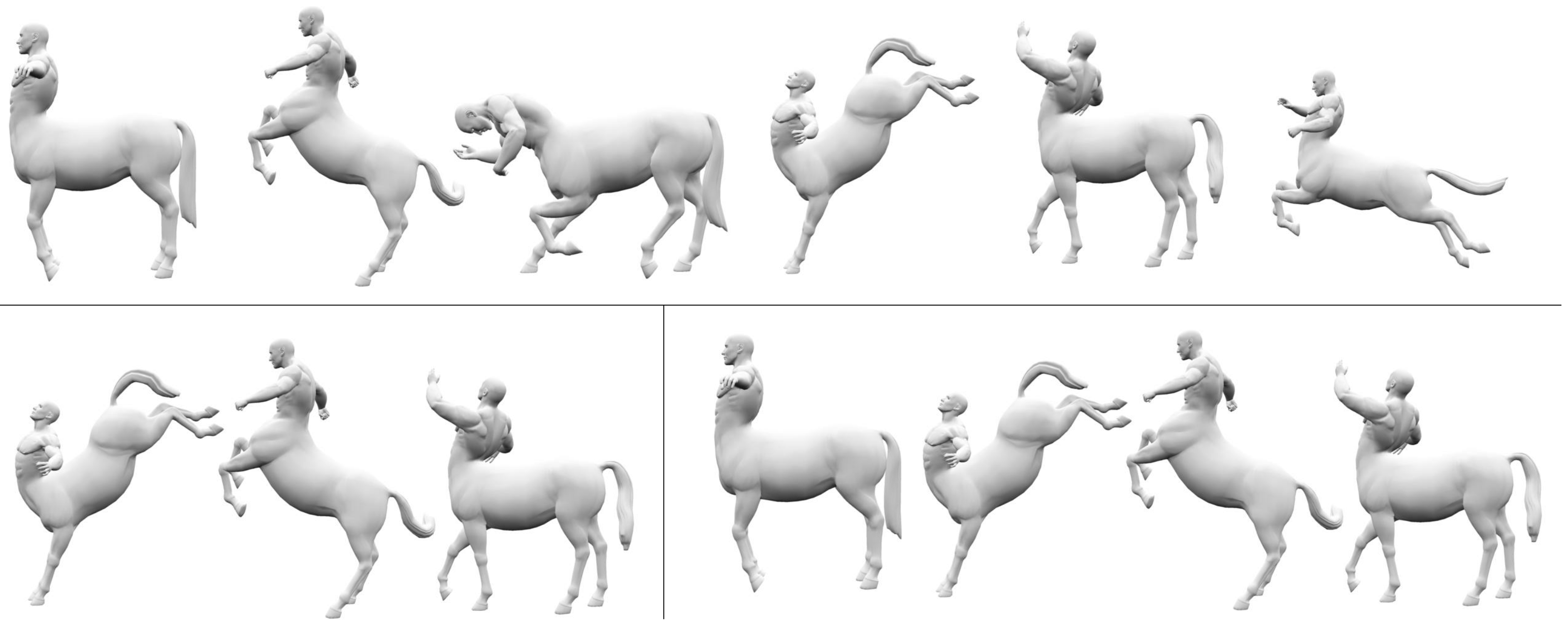}
    \caption{Centaurs-6 (top), Centaurs-3 (bottom left), Centaurs-4 (bottom right).}
    \label{fig:centaurs}
\end{figure}

\begin{figure}[h!]
    \centering
    \includegraphics[width=\columnwidth]{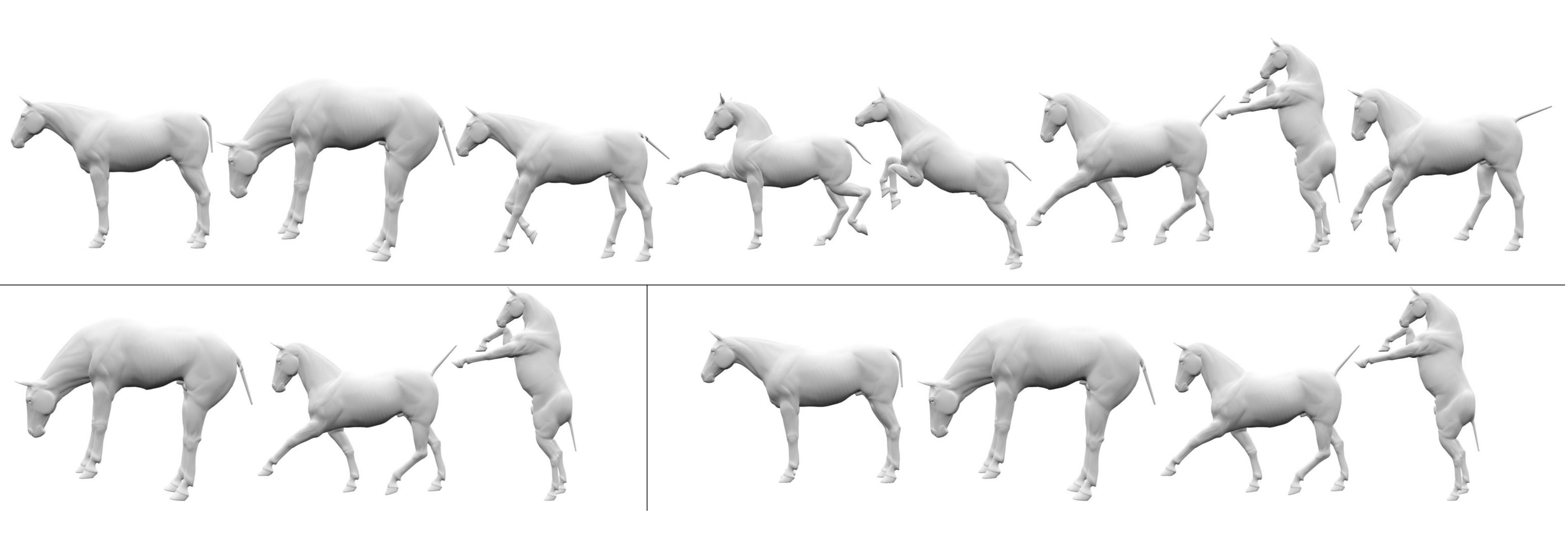}
    \caption{Horses-8 (top), Horses-3 (bottom left), Horses-4 (bottom right).}
    \label{fig:horses}
\end{figure}

\begin{figure}[h!]
    \centering
    \includegraphics[width=\columnwidth]{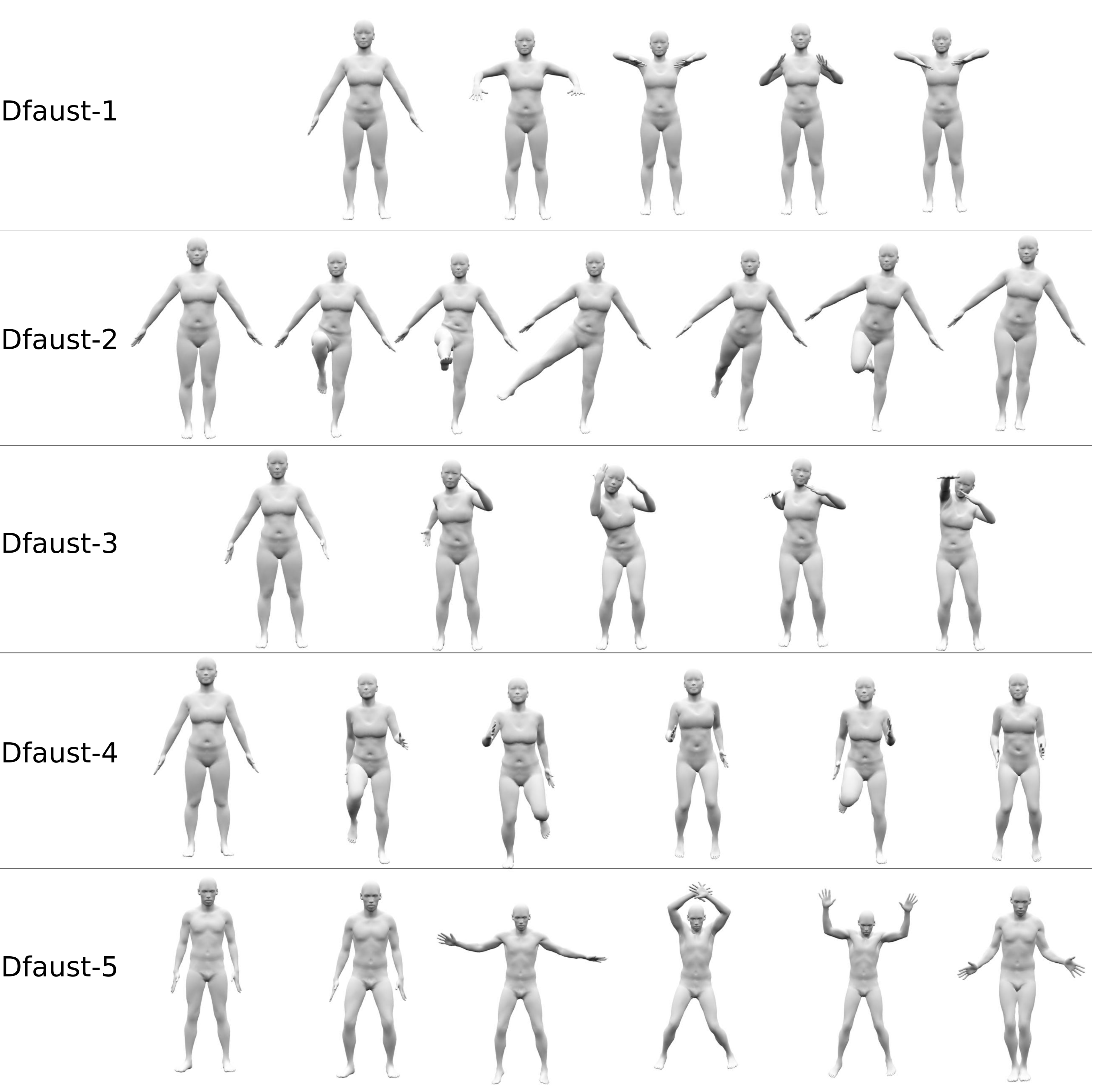}
    \caption{The keypoints of 5 different Dynamic Faust sequences used for training, for Table 2 of main paper.}
    \label{fig:faust_keypoints}
    \vspace{-\baselineskip}
\end{figure}

\newpage
\section{Network Architecture}
\begin{figure*}[t!]
    \centering
    \includegraphics[width=\linewidth]{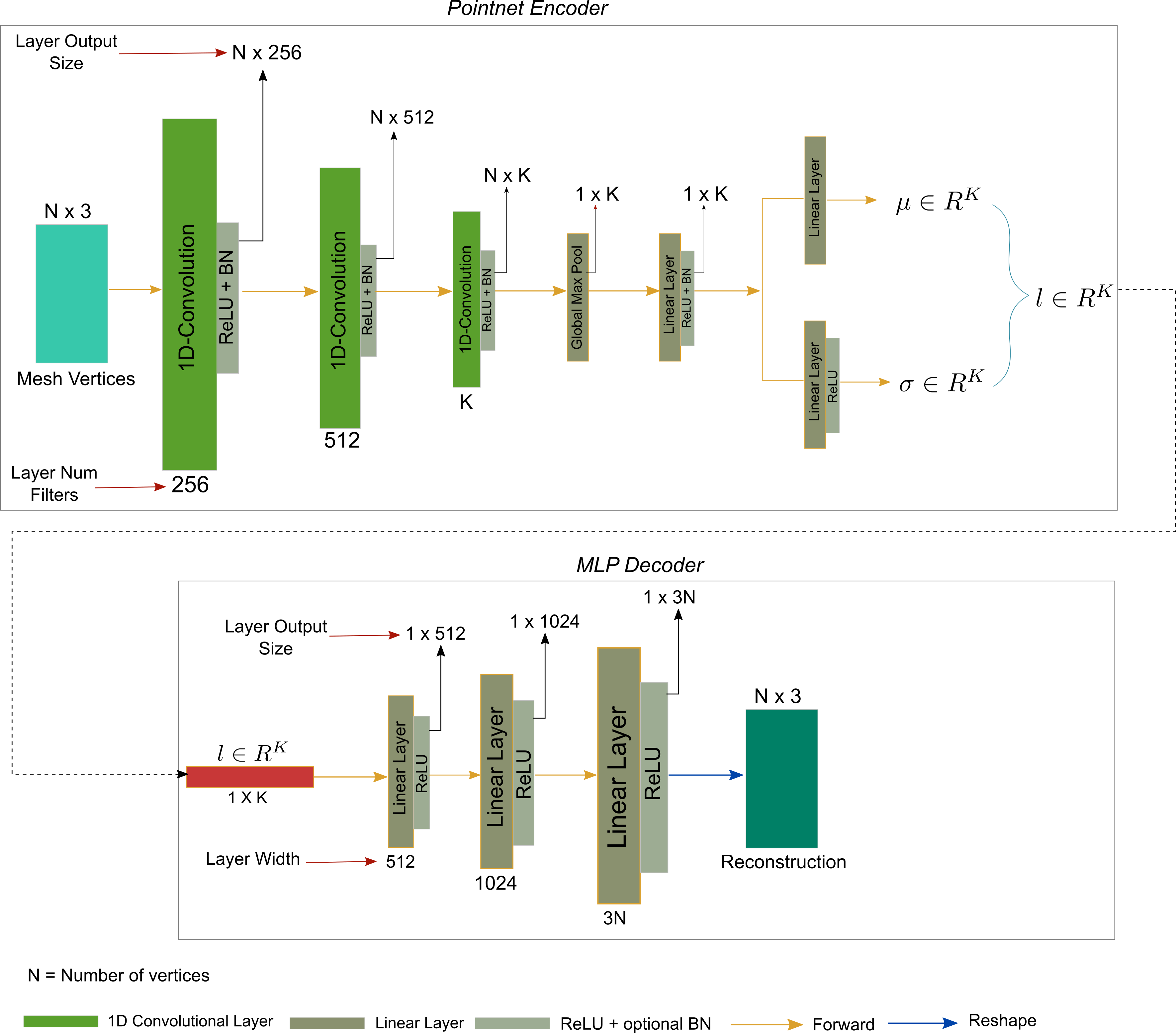}
    \caption{The VAE architecture used by GLASS.}
    \label{fig:network}
\end{figure*}

Figure~\ref{fig:network} shows the VAE architecture used by \name. The main pseudocode for \name is provided in the main paper. 

\section{Extrapolation}
We extrapolate beyond the endpoints of interpolation for the same length as interpolation and measure the smoothness of the mesh, and report the results in table \ref{tab:extrapolation}. We observe that extrapolated meshes have a smoother surface for \name than the baselines.

\begin{table*}[h!]
\label{tab:extrapolation}
\small
\centering
  \caption{Ablation study. Surface smoothness of extrapolated shapes. All results are normalized such that Vanilla VAE is 1.0, and lower numbers are better.}
  \label{tab:extrapolation}
  \begin{tabular*}{\textwidth}{r @{\extracolsep{\fill}} cccccc}
      ~Data~ & ~Vanilla VAE~ & ~ +Interpolation ~ & ~ +Interp. +Energy ~ &  ~LIMP~ & ~ +ARAP SI \cite{alexa2000arapinterp} ~ & \name\\
      \hline
      \hline
      Faust-3 &1.0 & 0.74 & 0.79 & 0.87 & 1.18 & \textbf{0.69}  \\
      Faust-5 & 1.0 & 0.64 & 0.6 & 0.63 & 1.07 & \textbf{0.59} \\
      Faust-7 & 1.0 & 0.66 & 0.66 & 0.64 & 0.94 & \textbf{0.63} \\
Faust-10 & 1.0 & 0.65 & \textbf{0.6} & 0.62 & 0.93 & \textbf{0.6} \\
\hline
Centaurs-3 & 1.0 & 1.13 & 1.03 & 1.11 & 1.21 & \textbf{0.79} \\
Centaurs-4 & 1.0 & 0.88 & 0.88 & 0.9 & 1.09 & \textbf{0.76}  \\
Centaurs-6 & 1.0 & 0.7 & \textbf{0.69} & 0.79 & 1.07 & \textbf{0.69} \\
\hline
Horses-3 & 1.0 & 1.1 & 1.06 & 1.1 & 1.22 & \textbf{0.75}  \\
Horses-4 & 1.0 & 0.87 & 0.82 & 0.8 & 1.09 & \textbf{0.71}  \\
Horses-8 & 1.0 & 0.62 & 0.62 & 0.61 & 1.0 & \textbf{0.59}  \\
    \end{tabular*}
\end{table*}

\end{document}